\documentclass{article}
\usepackage[final]{neurips_2024}

\usepackage[utf8]{inputenc} %
\usepackage[T1]{fontenc}    %
\usepackage{hyperref}       %
\usepackage{url}            %
\usepackage{booktabs}       %
\usepackage{amsfonts}       %
\usepackage{nicefrac}       %
\usepackage{microtype}      %
\usepackage[dvipsnames]{xcolor}
\usepackage{float}
\usepackage{amsmath}
\usepackage{mathtools}
\usepackage{csquotes}
\usepackage{graphicx}
\usepackage[capitalise,noabbrev]{cleveref}
\usepackage{tabularx}
\usepackage{array}
\usepackage{subcaption}
\usepackage{enumitem}
\usepackage{wrapfig}
\usepackage{placeins}
\usepackage{etoolbox}

\newcolumntype{Y}[1]{>{\raggedright\arraybackslash}m{#1\textwidth}}
\newcolumntype{C}{>{\centering\arraybackslash}X} %
\newcolumntype{L}{>{\raggedright\arraybackslash}X} %
\newcolumntype{M}[1]{>{\centering\arraybackslash}m{#1}} %

\newlength{\TItemSep}
\newlength{\TParSep}
\newenvironment{compactitemize}
 {\begin{itemize}[nosep,
                  leftmargin=*,
                  before=\vspace{-\TParSep+8pt},
                  after=\vspace{-\TParSep-5pt}]%
 }
 {\end{itemize}}

\setcitestyle{authoryear}

\newcommand{\nplink}[2]{%
  \href{https://www.neuronpedia.org/gpt2-small/#1-res-jb/#2}{\texttt{#1/#2}}%
}

\graphicspath{{figures/}}
\title{Evolution of SAE Features Across Layers in LLMs}

\author{%
  Daniel Balcells\footnotemark[1] \\
  Independent\\
  \And
  Benjamin Lerner\footnotemark[1] \\
  Independent\\
  \And
  Michael Oesterle\footnotemark[1] \\
  Independent\\
  \And
  Ediz Ucar\footnotemark[1] \\
  Independent\\
  \And
  Stefan Heimersheim\\
  Apollo Research\\
}

\makeatletter
\newif\iffinalorpreprint
\finalorpreprintfalse

\if@neuripsfinal
    \finalorpreprinttrue
\else
    \if@preprint
        \finalorpreprinttrue
    \fi
\fi
\makeatother

\begin{document}
\maketitle

\iffinalorpreprint
    \renewcommand{\thefootnote}{\fnsymbol{footnote}}
    \footnotetext[1]{Equal contribution.
    Correspondence to \href{mailto:stefan.heimersheim@gmail.com}{stefan.heimersheim@gmail.com}}
\fi

\begin{abstract}
    Sparse Autoencoders for transformer-based language models are typically defined independently per layer. In this work we analyze statistical relationships between features in adjacent layers to understand how features evolve through a forward pass. We provide a 
    \iffinalorpreprint
        \href{https://stefanhex.com/spar-2024/feature-browser/}{graph visualization interface}
    \else
        \href{http://vps-cde188ea.vps.ovh.net/}{graph visualization interface}
    \fi
    for features and their most similar next-layer neighbors, and build communities of related features across layers. We find that a considerable amount of features are passed through from a previous layer, some features can be expressed as quasi-boolean combinations of previous features, and some features become more specialized in later layers.
\end{abstract}

\section{Introduction}
\begin{figure}[htp]
    \centering
    \includegraphics[width=\textwidth]{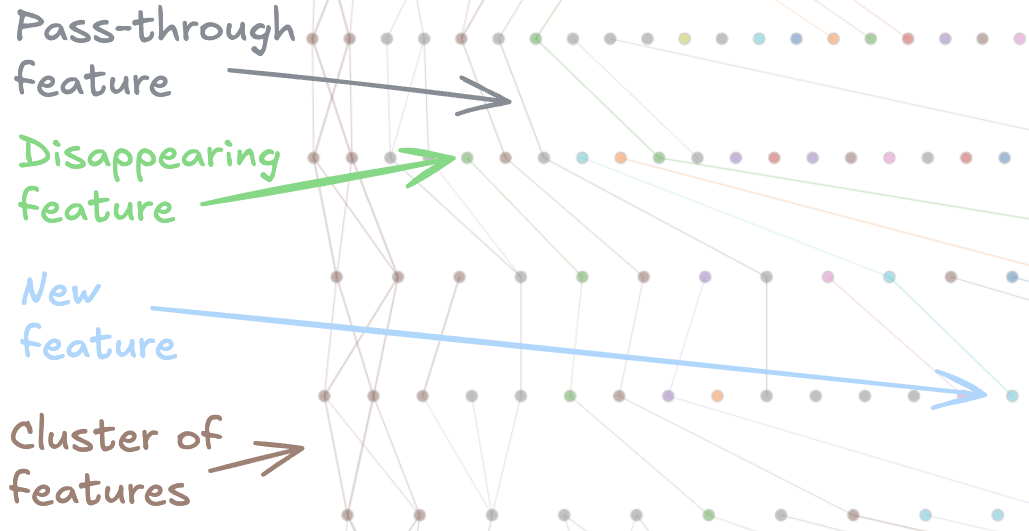}
        \caption{Different motifs we found in our feature graph, active on a single forward pass. Nodes represent SAE features in different layers, earlier layers are at the bottom. (a) \enquote{Pass-through} features have high correlation and similar semantic meaning between layers (b) \enquote{New} features don't have a counterpart in the preceding layer, we find some that appear 
    to be \texttt{AND}/\texttt{OR} gates of preceding features. (c) \enquote{Disappearing} features don't have a similar feature in the following layer. (d) We find many clusters of related features by running modularity detection algorithms on the full correlation graph, shown here as colors.}
    \label{fig:intro}
\end{figure}

The goal of Mechanistic Interpretability is to understand how neural networks implement algorithms and store information. The components that we use to define transformers, such as individual attention head neurons \citep{janiak-2023}, MLP neurons \citep{gurnee-2023}, or residual streams, do not seem to be a good basis for analysis, because they can represent multiple functions at once. SAEs \citep{sharkey-2022,cunningham-2023,bricken-2023} attempt to solve this problem by transforming uninterpretable neural network activations into a high-dimensional, overcomplete basis. These basis dimensions appear more monosemantic than neurons \citep{bricken-2023}.

We typically train and interpret SAEs and their features on each transformer layer independently. While features can be horizontally organised \citep[within a single SAE,][]{templeton-2024}, the features in different layers are not connected \enquote{vertically}. To understand SAE features, it would be helpful to know their upstream and downstream features. In the extreme (yet as we find, fairly common) case, features are simply duplicated across layers, likely corresponding to directions that are passed through the residual stream, but more complicated relationships are possible.

Related work \citep{marks-2024} computes causal relationships between SAE features active on a given prompt. Our research instead builds a graph of all SAE features, using cheaper correlational measures to find a more general (not prompt-specific, as advocated for in \citet{nanda-2022}) structure.

\begin{figure}[H]
  \centering
    \includegraphics[width=\textwidth]{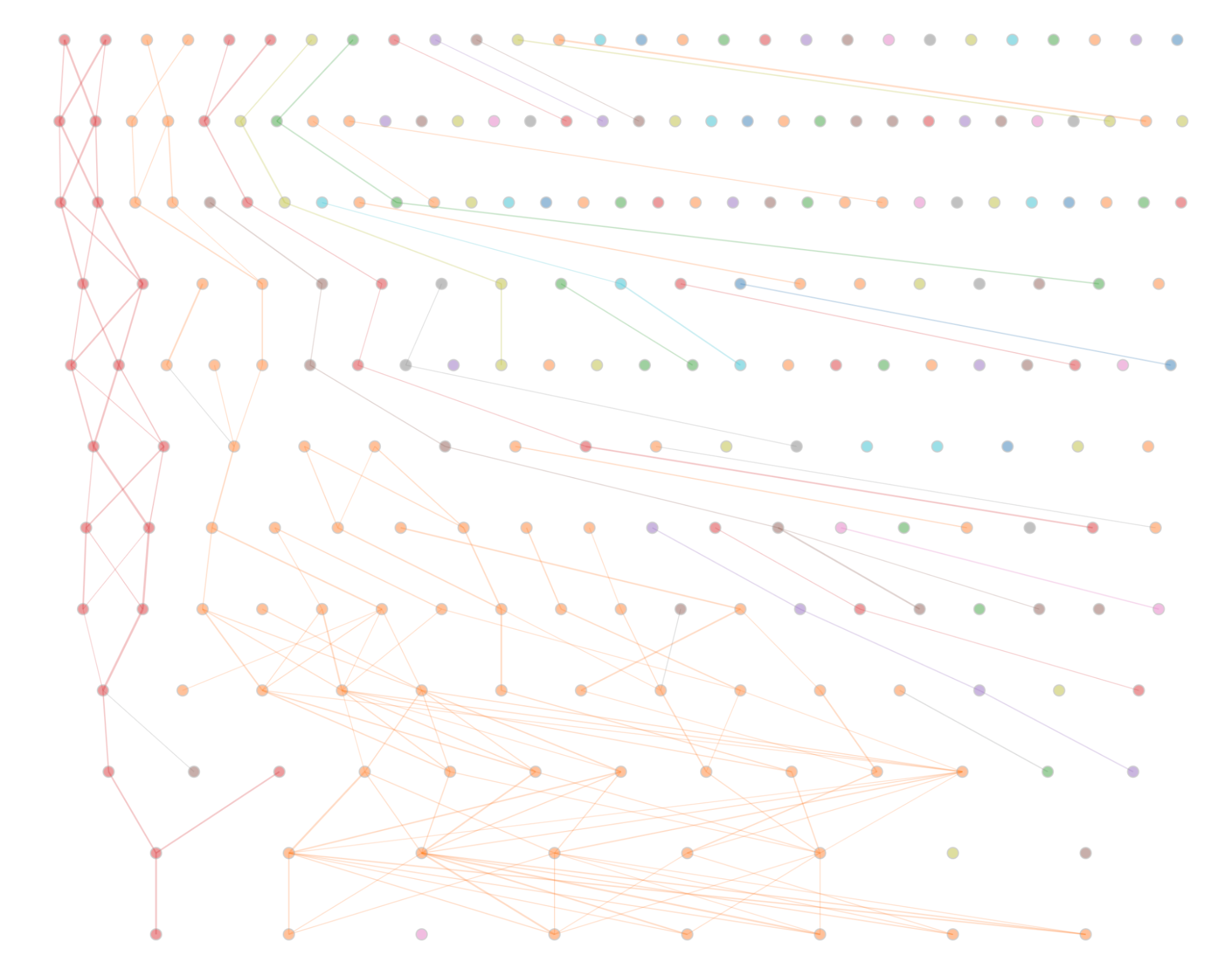}
    \caption{\textbf{The community structure within an SAE feature graph}. Nodes represent all the features that were active in the residual stream for a specific prompt of text. The rows of the graph correspond to layers of the transformer such that the bottom row corresponds with the first layer and the top row corresponds to the last layer. The edges between the nodes show the Jaccard similarity between two features $>0.1$. The nodes are coloured by the community they were assigned by the Leiden algorithm using the modularity quality function - nodes within a community are semantically similar to one another. For example, the pink community on the far left consists of features related to \enquote{Instructions directing on to take action}. This graph can be viewed in the feature browser by selecting the \texttt{jaccard\_leiden\_modularity\_threshold\_0.1\_masked\_single\_23} option in the settings.}
    \label{fig:graph-viz}
\end{figure}

\paragraph*{Contributions: }
We create a vertical feature graph, connecting SAE features across adjacent layers based on correlation measures (\cref{fig:graph-viz}), and provide 
\iffinalorpreprint
    \href{https://stefanhex.com/spar-2024/feature-browser/}{a web interface}
\else
    \href{http://vps-cde188ea.vps.ovh.net/}{a web interface}
\fi
to explore the feature graph. Qualitatively, we show that a sizable portion of features in a given layer are passed through from a previous layer (\cref{sec:results:pass-through}). We find instances in which multiple features in one layer frequently co-activate with a feature in the subsequent layer, in a manner resembling resembling \texttt{AND}/\texttt{OR} gates (\cref{sec:results:logic-gates}). We investigate subgraphs stretching across multiple layers composed of highly correlated features and show instances of features becoming more specialized in later layers (\cref{sec:results:communities}). We can spot instances in which GPT-3.5 incorrectly labeled features by noticing inconsistent explanations between neighboring features (\cref{app:autointerp-inaccuracies})---we expect this could improve automatic interpretability methods. We perform ablation checks and find that our correlational relationships often but not always correspond to causal relationships (\cref{app:ablation}).

\section{Methodology} \label{sec:methodology}
We leverage GPT-2-small \citep{radford2019language} and the \texttt{res-jb} SAEs  \citep{bloom-2024} trained on it. We construct multipartite network graphs with nodes representing SAE features, and edge weights being the similarity measure between two features in different layers. We then conduct network analysis on the resultant graph and explore its structure visually.

We denote SAE features as \texttt{L/F} with zero-based indices, e.g., feature 3465 in layer 4 is written as \nplink{4}{3465}. We leverage \href{https://www.neuronpedia.org/gpt2-small}{Neuronpedia}'s GPT-3.5 summary of the tokens that most highly activated each feature to annotate SAE features with explanations.

\paragraph{Feature similarity measures:} \label{sec:similarity-measures}
We collect SAE feature activations across all layers over 10M tokens in the Pile \citep[see \cref{app:pipeline}]{gao-2020} and then compute, for all adjacent-layer pairs of features: \textit{Pearson correlation}, \textit{Jaccard similarity} (how often both features fired as a fraction of the number of times at least one fired), \textit{Sufficiency} (how often the downstream feature fired when the upstream feature fired), and \textit{Necessity} (how often the upstream feature fired when the downstream feature fired). See \cref{app:max-activation-distribution} for details. We also considered other measures but found them to be less useful in early experiments (\cref{app:similarity-measures}).

\paragraph{Choosing subsets of nodes and edges for graph visualization:}
The full similarity graph of GPT-2-small for any given measure consists of $12 \cdot 24576 \approx 300$k nodes and $11 \cdot 24576^2 \approx 6$B edges. We perform post-processing (e.g. modularity calculations, see \cref{sec:detecting-communities}) on the full graph, but choose subsets for visualization purposes using either the set of features active on the final token in a given forward pass with a single prompt, or feature subsets based on community clustering algorithm (\cref{sec:results:communities}). We only visualize edges above a similarity threshold.

\paragraph{Detecting communities in SAE feature graphs:} \label{sec:detecting-communities}
We apply community detection algorithms in order to find structure in the graph. The Leiden algorithm \citep{traag-2019} can find well-connected communities of nodes - we predict that it would discover groups of semantically similar SAE features. We consider different methods of constructing the feature graph (see \cref{app:graph-construction}) before running the algorithm in order to yield different partitions of the graph.

\paragraph{Establishing causality between features through ablation:}
None of the similarity measures above can show that any SAE feature $f_k$ in layer $k$ \emph{causally} contributes to the magnitude of a feature $f_{k+1}$ in layer $k+1$. In order to demonstrate causality, we need to consider counterfactual situations---of all the inputs on which features $f_k$ and $f_{k+1}$ fire in the same forward pass, if $f_k$'s output were set to $0$ and everything else were kept constant, what would the average change in $f_{k+1}$'s output be? In \cref{app:ablation} we show that the Pearson correlation weakly correlates with the causal effect. We conclude that our feature graph often points out causal relationships, but should primarily be considered correlational.

\section{Results}

\subsection{Features being \enquote{passed through} multiple layers} \label{sec:results:pass-through}
\begin{figure}[htp]
    \centering
    \includegraphics[width=0.9\textwidth]{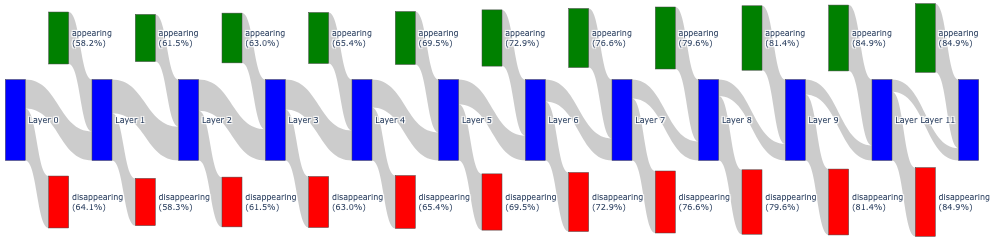}
    \caption{Passed-through/appearing/disappearing SAE features where Pearson $\geq 0.95$}
    \label{fig:pass-through}
\end{figure}

We say that a feature in one layer is \emph{passed through} if there is at least one very similar next-layer feature, that it \enquote{disappears} if it has no next-layer correspondent, and \enquote{appears} if it has no previous-layer correspondent.
\cref{fig:pass-through} illustrates the three classes of features per layer, for Pearson correlation
similarity with a threshold of $t=0.95$. We provide a discussion of different threshold choices in
\cref{app:pass-through}.
\cref{tbl:passthroughexamples} shows an example
of a passed-through feature throughout multiple layers (discovered with the community finding
method, see \cref{sec:results:communities}). We see that more than half of the features are appearing and disappearing at each layer, and this fraction increases to 80\% for the later layers. This could indicate that earlier layers \enquote{build up} features for later use, after which they are dropped from the residual stream.

\begin{table}[!ht]
    \begin{tabularx}{\textwidth}{M{1.5cm}M{3cm}L}
        \toprule
        \textbf{Feature} & \textbf{Pearson similarity to next layer feature} & \textbf{Neuronpedia explanation} \\
        \midrule
        \nplink{0}{21891} & 0.97 & Terms related to panic and anxiety \\
        \nplink{1}{19235} & 0.99 & Phrases related to feelings of intense fear or alarm \\
        \nplink{2}{20272} & 0.99 & instances of the word \enquote{panic} \\
        \nplink{3}{18762} & 0.99 & Words related to feelings of fear or distress \\
        \nplink{4}{17895} & 0.98 & Words related to a sense of urgency or crisis \\
        \nplink{5}{21017} & 0.96 & Words related to the concept of panic \\
        \bottomrule
    \end{tabularx}
    \caption{Example of a feature being passed through multiple layers. Features denoted as {layer/feature}.}
    \label{tbl:passthroughexamples}
\end{table}

\subsection{Logic Gates}\label{sec:results:logic-gates}
Necessity and sufficiency approximate logical implications if the similarity values are sufficiently close to 1: If, e.g., Necessity$(f_u, f_d) = 0.99$, then it is \enquote{99\% necessary} that the upstream feature $f_u$ is active for the downstream feature $f_d$ to be active; for two upstream features with this value, it is still \enquote{98\% necessary} that they both fired. Using these two measures, we can find approximations of \texttt{AND} and \texttt{OR} relationships between SAE features by searching for downstream features with multiple (but few) upstream neighbours with very high similarity scores. \cref{tbl:logic-gates} shows some features with exactly two neighbours; more examples and a comparison with the other two measures can be found in \cref{app:logic-gates}.

\begin{table}[htp]
    \begin{tabularx}{\textwidth}{Y{0.1}Y{0.58}Y{0.225}}
    \toprule
    \textbf{Measure} & \textbf{Upstream features} & \textbf{Downstream feature} \\
    \midrule
    
    Necessity (\texttt{AND}) &
    \begin{compactitemize}
        \item \nplink{1}{9238} (locations and countries in the Baltic region) 
        \item \nplink{1}{12192} (names of places, especially focusing on Estonia and individuals related to legal matters)
    \end{compactitemize} &
    \nplink{2}{20513} (mentions of the country Estonia) \\
        
    Necessity (\texttt{AND}) & 
    \begin{compactitemize}
        \item \nplink{1}{1411} (references to animals, specifically sheep) 
        \item \nplink{1}{13700} (the word \enquote{Lamb} preceded or followed by a specific suffix)
    \end{compactitemize} &
    \nplink{2}{1218} (mentions of or references to lamb) \\

    Sufficiency (\texttt{OR}) &
    \begin{compactitemize}
        \item \nplink{0}{303} (references to various organizations or entities with \enquote{Life} in their name) 
        \item \nplink{0}{7012} (terms related to MetLife Stadium, Half-Life)
    \end{compactitemize} &
    \nplink{1}{17435} (references to the term \enquote{Life}) \\
    
    Sufficiency (\texttt{OR}) &
    \begin{compactitemize}
        \item \nplink{1}{1208} (related to current status or ongoing actions) 
        \item \nplink{1}{6032} (text referring to current states or occurrences)
    \end{compactitemize} &
    \nplink{2}{4672} (instances where the term \enquote{currently} is used) \\
    \bottomrule
    \end{tabularx}

    \caption{Feature connections resembling logic gates, where necessity or sufficiency was $\geq0.999$}
    \label{tbl:logic-gates}
\end{table}

\subsection{Did features disappear, or did they lack representation in the next layer's SAE?} \label{sec:results:disappear}

\begin{figure}[htbp]
    \centering
        \includegraphics[width=5cm]{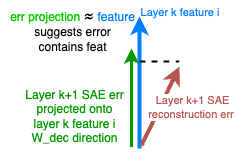}
        \hfill
        \includegraphics[width=8.7cm]{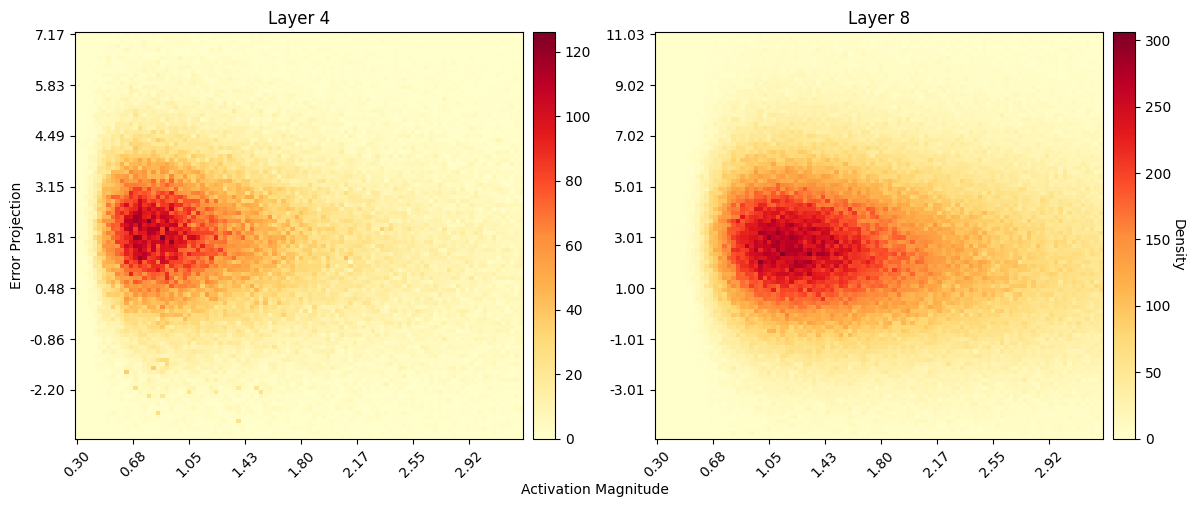}
    \caption{
    \textbf{Left:} Visualizing the recovery of the previous layer's features from the next layer's error term.
    \textbf{Right:} Heatmaps of SAE feature activation vs next-layer error projected onto the feature directions of the previous layer, across many tokens. Only features which \enquote{disappeared} (necessity $< 0.4$ with all next layer features), with an activation $>0.1\%$ of their max activation are shown.}
    \label{fig:err-projections}
\end{figure}

A feature having no meaningful similarity to any downstream SAE feature could indicate that that feature is no longer present in the downstream layer, or that it is not represented by the SAE but present in its reconstruction error
\citep{marks-2024}.
To understand the degree to which this happens, we compare the magnitude of an SAE feature $i$ in layer $k$ to the magnitude of layer $k+1$'s error term ($x_{\rm{resid}} - {x_{\rm{resid}}W_{\rm{enc}}W_{\rm{dec}}}$) projected onto $W_{\rm{dec}}^{k}[i]$, as illustrated in \cref{fig:err-projections}. We provide a detailed description of this method in
\cref{app:projectionmethod}.

We do not see evidence that disappearing features are present in the next layer's residual stream. This suggests that these features no longer exist downstream, possibly because they split or merge with other features.

\subsection{Community-detection finds semantically meaningful subgraphs} \label{sec:results:communities}
\begin{figure}[htbp]
    \centering
        \includegraphics[height=3.5cm]{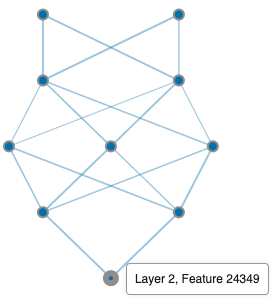}
        \hfill
        {\color{lightgray}\vrule width 0.5pt}
        \hfill
        \includegraphics[height=3.5cm]{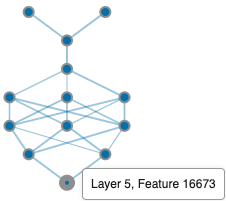}
        \hfill
        {\color{lightgray}\vrule width 0.5pt}
        \hfill
        \includegraphics[height=3.5cm]{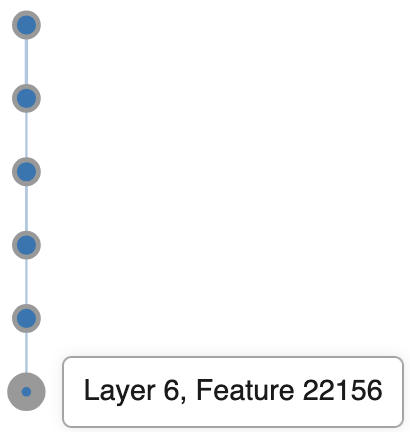}
        \caption{Communities found by applying the Leiden algorithm with a modularity quality function. Intra-layer feature cosine similarity in each community (not used for forming the communities) was measured to be $\ge0.75$.
    \textbf{\href{https://www.neuronpedia.org/list/cm158ljma000dbz0hs79d5obl}{Left}}: Necessity based. \nplink{2}{24349} detects \enquote{evidence} in the context of court cases, causes of economic phenomena, and scientific data. Downstream, \nplink{4}{8314} specializes in court evidence.
    \textbf{\href{https://www.neuronpedia.org/list/cm158pj0m0001v04u1xl5o82y}{Center}}: Necessity based. \nplink{5}{16673} detects \enquote{special} in several contexts, \nplink{7}{5871} focuses on \enquote{Special} in the title of a law enforcement official.
    \textbf{\href{https://www.neuronpedia.org/list/cm158t2jc0001127a5v8k7bco}{Right}}: Jaccard based. \nplink{6}{22156}, and each feature downstream detects the concept of \enquote{an important moment}.}
    \label{fig:combined-communities}
\end{figure}

We used both the Louvain and Leiden community detection algorithms (details in \cref{app:graph-construction}) to discover subgraphs within the similarity measure graphs. We observe in \cref{fig:combined-communities} (more examples in \cref{app:graph-construction}) a variety of structures within each similarity measures' communities. Simple communities are often long chains of pass-through features, one per layer, sharing many top activating tokens. In more complex communities, features within the same layer have high cosine similarity and activate on semantically similar tokens. In this way, individual layers in each community have similar properties to the clusters discovered in \citet{bricken-2023}. In some communities (\cref{fig:combined-communities}), we can trace \enquote{general} upstream features which detect a token in several contexts to features which appear to  \enquote{specialize} in detecting that same token in mutually exclusive contexts.

\section{Discussion}
Our approach measures correlation between features, not causation. This means we do not prove that features interact or that one feature affects the other; in principle, two correlated features could fire via completely unrelated mechanisms. However, it seems more plausible that the downstream feature fires because of the upstream feature (and we see some evidence of that in our ablation results in \cref{app:ablation}).

The sparsity of SAE feature activations generally results in a low number of active samples, especially since we are interested in co-activations, i.e., tokens where both features of a pair are firing (see \cref{app:activation-sampling} and particularly \cref{fig:co-activation} for details about the number of co-activating feature pairs). If only a few (say, less than 10) co-activations exist between feature pairs, the identified correlations might be spurious. Run-time constraints do not allow us to use datasets with billions of tokens to avoid this situation. On the other hand, \cref{fig:10-vs-100} shows that the effect of using 100M instead of 10M tokens has a negligible impact on the similarity matrices. 

We chose to focus on four specific similarity measures due to their ease of implementation and widespread use in statistics. However, many other similarity measures exist, and results might differ for other measures.

In addition to our correlation measures, we considered feature geometry: we briefly investigated cosine similarity between features (like \citet{templeton-2024}, but for different layers) but found it less useful on its own. A combination of both statistical and geometric measures might further improve the results, but is out of scope for this work.

The choice of hyperparameters---in our case, activation and similarity thresholds---is crucial for managing the wealth of raw data and creating interpretable results and visualisations. 

The community detection algorithms used are general purpose. The application of more specialised community-detection algorithms that are more appropriate for sparse multipartite graphs could find more semantically valid communities. A possible strategy for this is the use of a custom quality function for the Leiden algorithm that applies a penalty to the number of nodes used in the same layer.
We mostly rely on GPT3.5's summary of the top activating tokens of each SAE feature to understand what it is doing, even though we see in \cref{app:autointerp-inaccuracies} that these can be inaccurate.

We do not have a theory describing how feature specialization occurs. More work is needed to determine how a \enquote{specific} context is represented mathematically, e.g. comparing the SAE decoder weights for a \enquote{general} upstream feature to a \enquote{specific} downstream feature and trying to find where the diff lies within the weights of the network.

\section{Conclusion}
We have shown that SAE features in different layers have clear relationships, and created a graph to represent these relationships. \iffinalorpreprint
    \href{https://stefanhex.com/spar-2024/feature-browser/}{Our visualization is available (here).}
\else
    \href{http://vps-cde188ea.vps.ovh.net/}{Our visualization is available here.}
\fi

We identified communities of features in the graph which have semantically similar interpretations. These include connections resembling \texttt{AND} and \texttt{OR}-gates, examples of which may help us understand how features are computed in models.
Other communities were features being simply {passed-through} from layer to layer, likely via the residual stream. These features occupy entries of the SAE dictionary without contributing new information. We expect that this insight could be used to create more efficient SAEs across multiple layers.
Finally we use the graph to notice connected features with apparently-unrelated explanations, allowing us to spot incorrect (GPT3.5-generated) autointerp explanations. We hope that these correlated features could enrich or verify future autointerp methods.

\makeatletter
\iffinalorpreprint
\section*{Author Contributions}
\textbf{Daniel Balcells} performed implementation of the batched architecture for similarity calculation, implementation and optimization of Jaccard similarity and decoder cosine similarity measures; design and implementation of the ablation pipeline; design and implementation of the feature browser.
\textbf{Benjamin Lerner} conducted literature review, implemented ablation pipeline, projected error terms onto feature directions, analyzed communities for structure and meaning, drove the writing and editing of the paper.
\textbf{Michael Oesterle} designed and implemented the batched architecture for similarity computation, ran experiments to create and compress the matrices, computed statistics (max activations, co-activation of features, distribution of similarity values, pass-through features, ...), implemented the creation and annotation of graphs from prompts, analyzed graphs for structure and meaning, and contributed to writing the paper.
\textbf{Ediz Ucar} implemented the batched computation of the mutual information similarity measure, conducted an exploratory review of community detection methodologies, implemented algorithm agnostic community detection pipeline, ran experiments using similarity measures to cluster SAE graphs into communities using a broad spectrum of methods, algorithms and parameterisations, and contributed to writing and editing the paper. 
\textbf{Stefan Heimersheim} supervised the project and provided high-level guidance and ideas.

\section*{Acknowledgments}
This research was supported by the Center for AI Safety Compute Cluster. Any opinions, findings, and conclusions or recommendations expressed in this material are those of the authors and do not necessarily reflect the views of the sponsors.

We would like to thank Josh Purtell for giving feedback on the initial idea and providing helpful references to related literature.
\else
\fi
\bibliographystyle{plainnat}
\bibliography{bibliography}

\newpage \appendix

\section{Computing feature similarity} \label{app:pipeline}
Starting from input tokens, we extract the model activations from the \texttt{hook\_resid\_pre} residual stream position of each layer and feed them through the corresponding SAE. The resulting SAE activation vectors are scored pairwise using the similarity measures defined in \cref{sec:similarity-measures}.

\begin{figure}[htp]
    \centering
    \includegraphics[width=\textwidth]{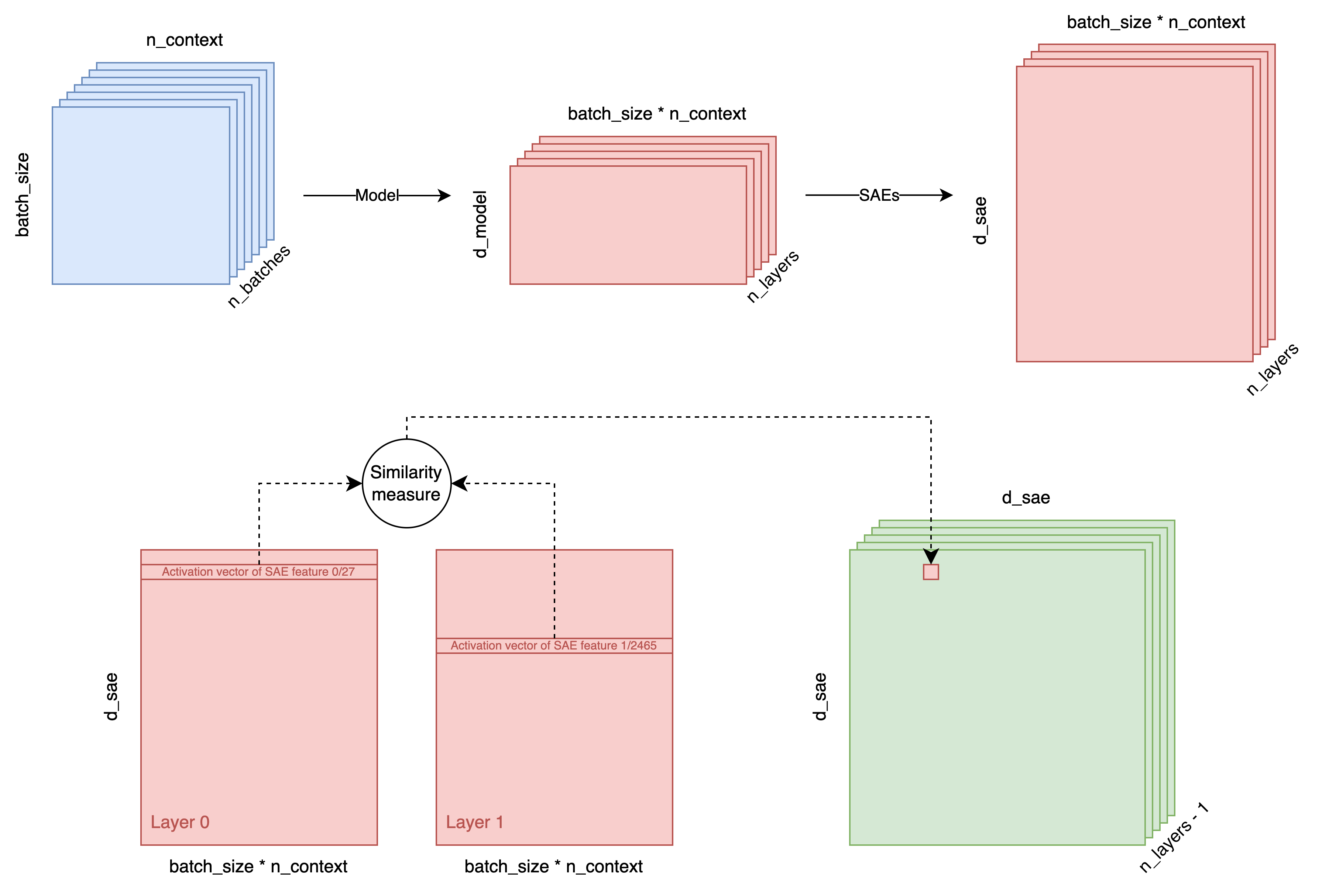}
    \caption{SAE feature similarity computation pipeline from tokens (blue) to model/SAE activations (red) to pairwise feature similarities (green).}
    \label{fig:pipeline}
\end{figure}

\section{Distribution of maximum activation values per feature} \label{app:max-activation-distribution}
\begin{wrapfigure}{r}{0.6\textwidth}
    \centering
    \includegraphics[width=0.6\textwidth]{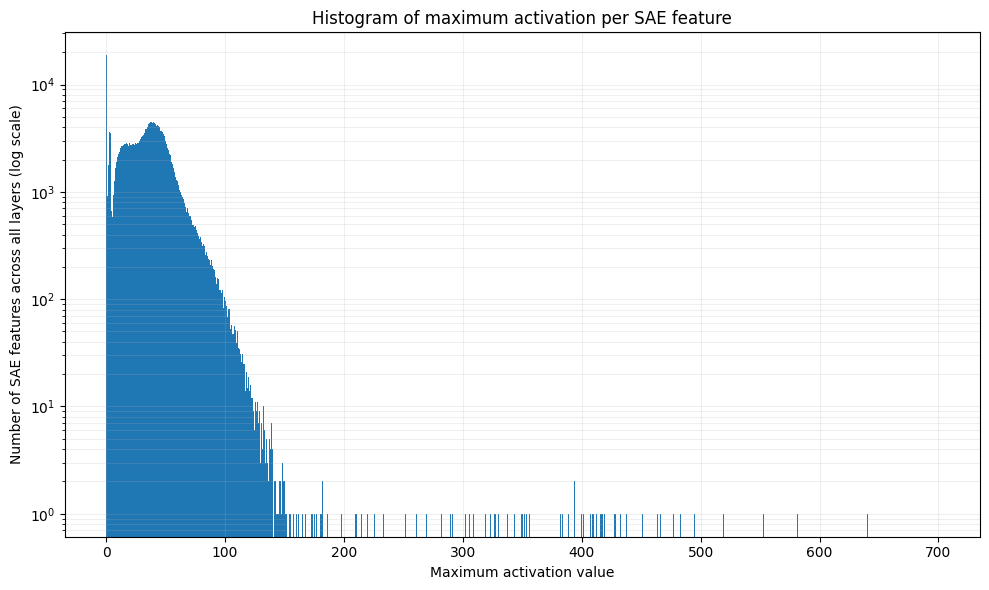}
    \caption{Maximum activation values of all SAE features over the 17.5M token dataset.}
    \label{fig:max-activation}
\end{wrapfigure} 

Whether or not an SAE feature is seen as \enquote{active} is determined by its (non-negative) activation value. However, the maximum activation values of each SAE over the dataset vary widely (see \cref{fig:max-activation}). Therefore, absolute activation thresholds---that is, defining a feature as active if its current activation values exceed a constant value---do not seem to be plausible since this value might never be exceeded by feature $f$, but almost always for feature $f'$. Therefore, we use relative activation thresholds throughout the paper, meaning that the threshold value is chosen between 0 and 1, and each feature's activation value is normalised by dividing it by its maximum activation (over the 17.5M token dataset) to determine whether the feature is active. Before computing Jaccard, sufficiency, and necessity we first binarize each feature by defining it as \enquote{active} if its activation is at least \textbf{20\%} of its maximum.

\section{Feature Activation Sampling} \label{app:activation-sampling}
\begin{wrapfigure}{r}{0.5\textwidth}
    \centering
    \includegraphics[width=0.5\textwidth]{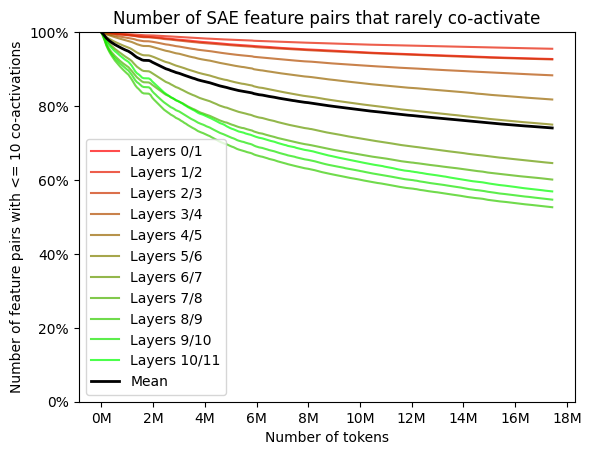}
    \caption{Number of SAE feature pairs from adjacent layers that rarely co-activate.}
    \label{fig:co-activation}
\end{wrapfigure}

Across the 11 pairs of adjacent layers, we measure the percentage of all $\sim6$B feature pairs that don't fire together (to avoid spurious correlations, we define \enquote{never} as \enquote{10 or fewer co-activations}) for datasets ranging in size between 500k and 20M tokens. We observe, in accordance with our expectation, that for small datasets, feature sparsity leads to a large proportion of feature pairs not co-activating (see \cref{fig:co-activation}). For example, for layers 4-5, using 2M tokens resulted in at least 96\% of feature pairs never firing together, compared with only 86\% of feature pairs for 10M tokens. Frequent co-activation points to a connection between features, but we need a high number of tokens to obtain a representative sample of co-activations to reliably compute similarities.

\begin{figure}[htp]
    \begin{subfigure}{0.5\textwidth}
        \includegraphics[width=0.9\linewidth, height=6cm]{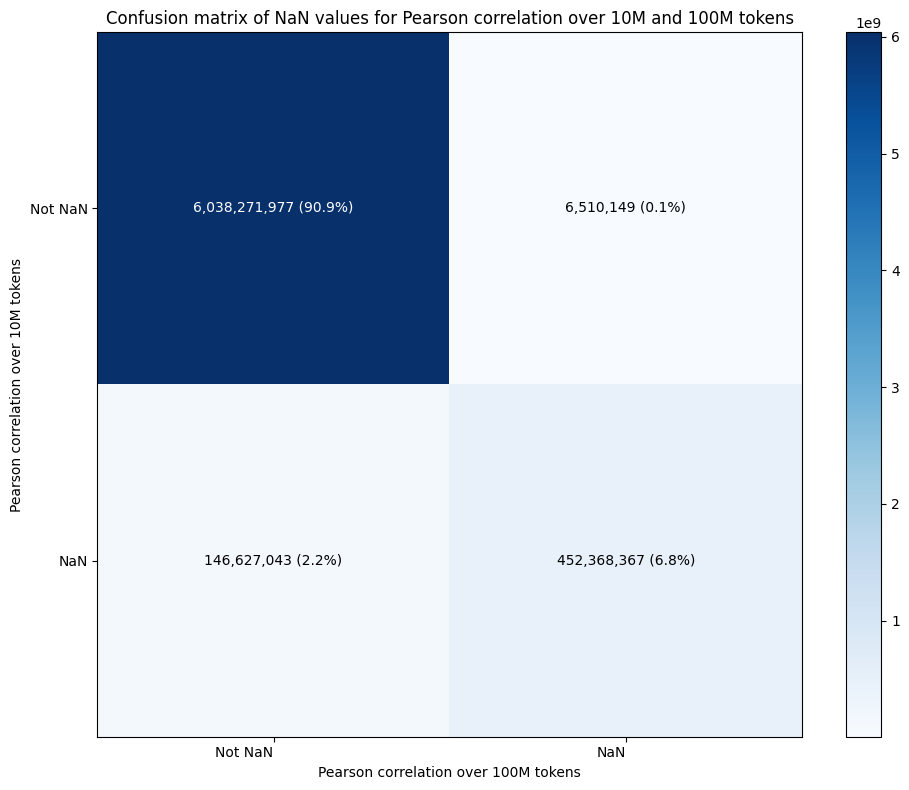} 
        \caption{Confusion matrix of \emph{NaN} entries.}
        \label{fig:10-vs-100-confusion}
    \end{subfigure}
    \begin{subfigure}{0.5\textwidth}
        \includegraphics[width=0.9\linewidth, height=6cm]{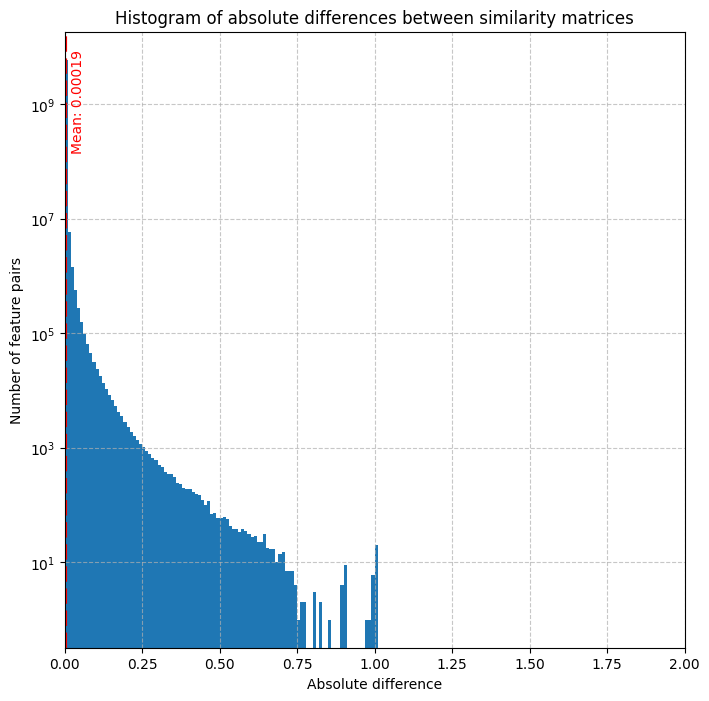}
        \caption{Histogram of absolute differences in similarity values.}
        \label{fig:10-vs-100-difference}
    \end{subfigure}
    \caption{Comparing the Pearson similarity matrices for 10M and 100M tokens.}
    \label{fig:10-vs-100}
\end{figure}

To verify that the improvement after 10M tokens is acceptably low, we compute the Pearson matrix for 100M tokens and compare its entries to the 10M token version (\cref{fig:10-vs-100}). We find a 97.7\% overlap in nan entries—those entries indicate that there was no co-activation of the respective feature pair in the dataset), and the mean absolute difference between corresponding entries is 0.00019.

Based on this data, we decide to use 10M tokens for all our similarity measures.

It is not feasible to calculate the similarity matrices after caching all activations since there are 3 trillion activations in total. Instead, for a given pair of layers, we run a batch of $32 \times 128$ tokens through the model, cache the SAE features, and compute and accumulate statistics (e.g. counts, sums and sums of squares for Pearson) of the batch. Once all batches have been processed, we finalise the aggregation (e.g. computing Pearson from the final means, variances and standard deviations).

\section{Distribution of similarity values}
\begin{figure}[htp]
    \begin{subfigure}{0.5\textwidth}
        \includegraphics[width=0.9\linewidth]{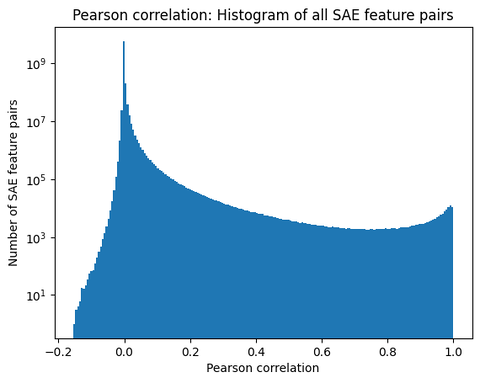} 
        \label{fig:pearson-histogram}
    \end{subfigure}
    \begin{subfigure}{0.5\textwidth}
        \includegraphics[width=0.9\linewidth]{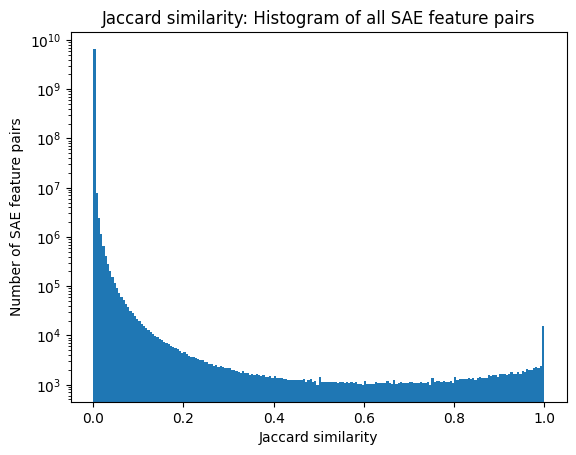} 
        \label{fig:jaccard-histogram}
    \end{subfigure}
    \begin{subfigure}{0.5\textwidth}
        \includegraphics[width=0.9\linewidth]{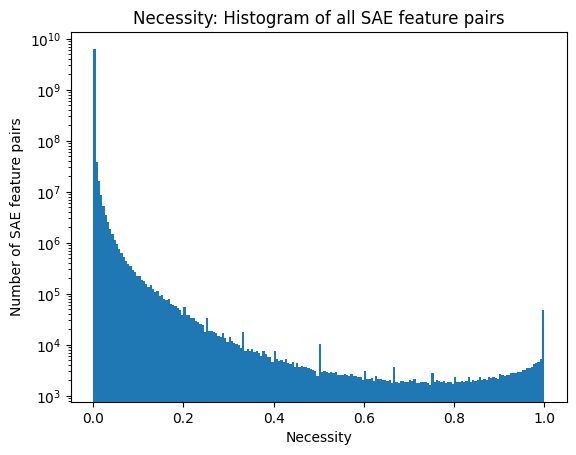} 
        \label{fig:necessity-histogram}
    \end{subfigure}
    \begin{subfigure}{0.5\textwidth}
        \includegraphics[width=0.9\linewidth]{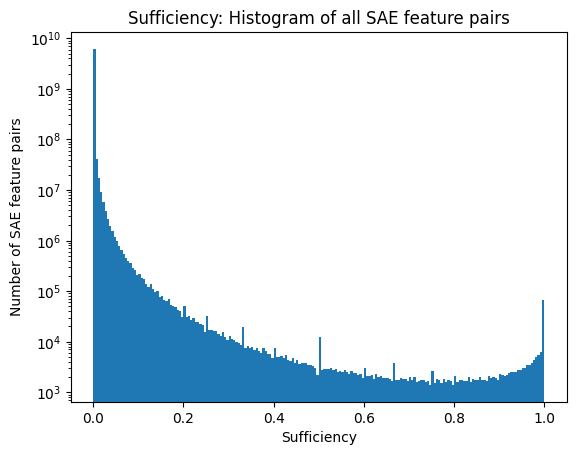} 
        \label{fig:sufficiency-histogram}
    \end{subfigure}
    \caption{Histograms (for all layer pairs) for the four similarity measures defined in \cref{sec:similarity-measures}. Each plot is based on 6B values minus the \emph{NaN} entries produced by non-co-activating feature pairs.}
    \label{fig:similarity-histograms}
\end{figure}

The approach described above provides us with a similarity measure for each of the metrics we use, for each pair of consecutive layers, and for each feature pair. In total, this gives us a tensor of shape \texttt{(n\_metrics, n\_layers - 1, n\_features, n\_features)}$ = (4, 11, 24576, 24576)$, which is $\approx 100$ GB of raw data. Many similarity values in this matrix are close to zero and, therefore, not likely to describe a strong connection between features. \cref{fig:similarity-histograms} shows the histograms of similarity values, one plot per measure (note the log scale of the y-axis):

Since the similarity values close to zero are not used for downstream processing, we replace all similarity scores below $0.1$ with $0$.

\section{Additional similarity measures} \label{app:similarity-measures}
A fast and easy method to measure the connection between SAE features is to simply compute the pairwise cosine similarity of their decoder weight vectors. The intuition is that, since there are residual connections between the residual streams of subsequent layers, we expect the structure of the RS spaces to be similar. Thus, the directions of SAE features in subsequent layers are more aligned (i.e., their cosine similarity is higher) when their interpretations are more similar. 
On the other hand, this approach has two weaknesses: First, the internal organisation of the residual streams might slowly (or even disruptively?) change over the course of multiple layers, such that the similarity of feature semantics is not properly represented by the similarity of their directions. Second, this analysis is \enquote{static} insofar as it does not take into account the data passed through the model. Early results with this measure were not promising, so we chose not to analyze it further.

\begin{figure}[htp]
    \begin{subfigure}{0.5\textwidth}
        \includegraphics[width=0.9\linewidth, height=6cm]{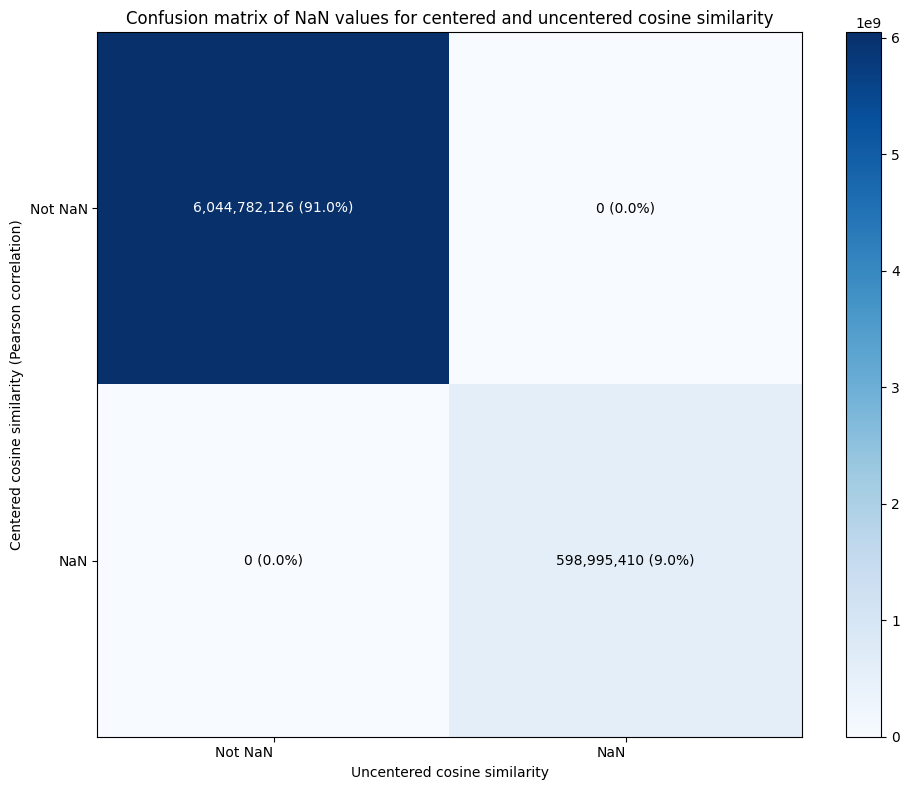} 
        \caption{Confusion matrix of \emph{NaN} entries.}
        \label{fig:centered-vs-uncentered-confusion}
    \end{subfigure}
    \begin{subfigure}{0.5\textwidth}
        \includegraphics[width=0.9\linewidth, height=6cm]{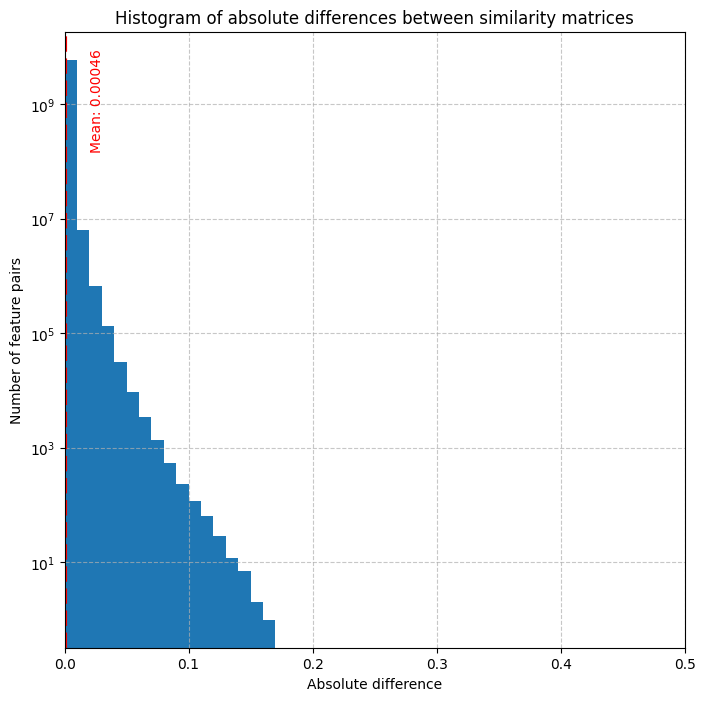}
        \caption{Histogram of absolute differences in similarity values.}
        \label{fig:centered-vs-uncentered-difference}
    \end{subfigure}
    \caption{Comparing the similarity matrices for centered (i.e., 
    Pearson) and uncentered cosine similarity.}
    \label{fig:centered-vs-uncentered}
\end{figure}

One additional statistical measure we considered is an uncentered Pearson correlation ${\rm Similarity}(x,y)=1/N \sum_{i=1}^N x_i y_i$ for $N$ samples. This is essentially the \enquote{cosine similarity} between the activation sample, but we caution against that name as it has nothing to do with the geometric cosine similarity mentioned above. \cref{fig:centered-vs-uncentered} compares the centered and uncentered Pearson correlation.

\section{Explanation pairs for different similarity values} \label{app:explanation-pairs}
To build a better understanding about the connection between similarity scores and intuitive \enquote{closeness} of textual explanations, \cref{tbl:pearson-explanations,tbl:jaccard-explanations,tbl:necessity-explanations,tbl:sufficiency-explanations} list randomly sampled explanation pairs for a wide range of similarity scores, using all of our measures. Note that the GPT-3.5 explanations are often inaccurate, and checking the top activating tokens on Neuronpedia might provide more insight than just reading the explanations.

\begin{table}[htp]
	\begin{tabularx}{\textwidth}{Y{0.1}Y{0.4}Y{0.4}}
	\toprule
	\textbf{Similarity} & \textbf{Upstream feature} & \textbf{Downstream feature} \\
	\midrule
		-0.000 & \nplink{3}{6743} (phrases related to rebranding or marketing strategies) & \nplink{4}{21451} (words related to being interested or showing interest in something) \\ \cmidrule(lr){1-3}
		0.000 & \nplink{9}{849} (positive reviews or feedback on written works) & \nplink{10}{4028} (Phrases related to events leading up to a significant moment) \\ \cmidrule(lr){1-3}
		0.000 & \nplink{2}{3681} (proper nouns referring to people or characters) & \nplink{3}{11698} (terms related to jars or containers) \\ \cmidrule(lr){1-3}
		0.100 & \nplink{10}{13029} (terms related to laws, rights, and regulations) & \nplink{11}{15564} (words related to general commentary or opinions) \\ \cmidrule(lr){1-3}
		0.136 & \nplink{10}{19189} (proper names, potentially female names) & \nplink{11}{19481} (geographic locations, particularly cities, states, and countries) \\ \cmidrule(lr){1-3}
		0.249 & \nplink{2}{4867} ( words related to negative attributes or actions) & \nplink{3}{11782} (phrases related to the concept of efforts or actions being futile or in vain) \\ \cmidrule(lr){1-3}
		0.208 & \nplink{10}{21191} (technology-related terms and concepts) & \nplink{11}{18002} (terms related to digital technologies, tools, and platforms) \\ \cmidrule(lr){1-3}
		0.357 & \nplink{9}{8132} (personal pronouns and forms of the verb \enquote{to be} associated with self-identification) & \nplink{10}{9550} (phrases related to direct speech and thoughts) \\ \cmidrule(lr){1-3}
		0.415 & \nplink{10}{24179} (words related to legal and political entities) & \nplink{11}{18344} (information related to individuals and their actions or controversies surrounding them) \\ \cmidrule(lr){1-3}
		0.433 & \nplink{10}{14658} (sources or credits attributions in text) & \nplink{11}{7285} (news sources or citations in a specific format) \\ \cmidrule(lr){1-3}
		0.597 & \nplink{6}{23635} (sensitive issues or controversy in a text) & \nplink{7}{13216} (MDA and ADA-related entities) \\ \cmidrule(lr){1-3}
		0.518 & \nplink{6}{4867} (conjunctions, specifically the word \enquote{and.}) & \nplink{7}{16193} (phrases indicating contrast or continuation) \\ \cmidrule(lr){1-3}
		0.659 & \nplink{4}{9929} (locations, organizations, and names related to military and government operations) & \nplink{5}{7308} (references to the hockey team \enquote{Montreal Canadiens.}) \\ \cmidrule(lr){1-3}
		0.646 & \nplink{5}{13549} (brands, names, and initials) & \nplink{6}{21248} (names starting with \enquote{Nik}) \\ \cmidrule(lr){1-3}
		0.757 & \nplink{4}{15057} (technical terms and jargon related to various fields or professions) & \nplink{5}{2506} (musical elements, such as instruments and music genres) \\ \cmidrule(lr){1-3}
		0.709 & \nplink{0}{3075} (the surname \enquote{Mul} with varying numbers following it) & \nplink{1}{23866} (terms related to the city of Varanasi) \\ \cmidrule(lr){1-3}
		0.830 & \nplink{9}{3936} (words related to geographical locations and organizations, specifically in the context of British Columbia (B.C.)) & \nplink{10}{23868} (locations mentioned along with abbreviated state names) \\ \cmidrule(lr){1-3}
		0.824 & \nplink{8}{9666} (long number sequences) & \nplink{9}{12911} (phone numbers) \\ \cmidrule(lr){1-3}
		0.976 & \nplink{10}{14867} (proper nouns of people's names) & \nplink{11}{8681} (the name \enquote{Don} in various contexts) \\ \cmidrule(lr){1-3}
		0.989 & \nplink{9}{1203} (phrases related to various aspects of development, including community development, software development, and personal development) & \nplink{10}{17959} (phrases related to personal or professional growth and progress) \\ 
		\bottomrule
	\end{tabularx}
	\caption{Explanation pairs for different values of Pearson}
	\label{tbl:pearson-explanations}
\end{table}

\begin{table}[htp]
	\begin{tabularx}{\textwidth}{Y{0.1}Y{0.4}Y{0.4}}
	\toprule
	\textbf{Similarity} & \textbf{Upstream feature} & \textbf{Downstream feature} \\
	\midrule
		0.000 & \nplink{6}{11361} (words related to finality or conclusion) & \nplink{7}{4252} (references to the iTunes Store) \\ \cmidrule(lr){1-3}
		0.000 & \nplink{3}{10208} (words related to different types of crusts) & \nplink{4}{5613} (references to the Islamic State group (ISIS)) \\ \cmidrule(lr){1-3}
		0.117 & \nplink{7}{22222} (phrases related to crime and violence against specific groups or individuals) & \nplink{8}{5232} (keywords related to legal actions, specifically criminal charges being filed against individuals) \\ \cmidrule(lr){1-3}
		0.132 & \nplink{5}{19423} (high-ranking government officials and officials in government positions) & \nplink{6}{23354} (names of political leaders or titles, possibly related to international relations) \\ \cmidrule(lr){1-3}
		0.209 & \nplink{6}{19522} (phrases related to inclusion, encompassing everything or everyone) & \nplink{7}{16587} (phrases emphasizing inclusivity and unity) \\ \cmidrule(lr){1-3}
		0.239 & \nplink{7}{852} (terms related to action genres, such as action movies and action-packed experiences) & \nplink{8}{18211} (words related to action and physical movement) \\ \cmidrule(lr){1-3}
		0.322 & \nplink{5}{16066} (phrases or words that are critical or judgmental in nature) & \nplink{6}{6754} (phrases related to comments or observations made by individuals) \\ \cmidrule(lr){1-3}
		0.363 & \nplink{9}{259} (statistics and trends over time) & \nplink{10}{15137} (phrases related to time and history) \\ \cmidrule(lr){1-3}
		0.478 & \nplink{7}{6833} (phrases related to time and duration) & \nplink{8}{15460} (durations of time in various units) \\ \cmidrule(lr){1-3}
		0.486 & \nplink{4}{5248} (words related to seriousness or urgency) & \nplink{5}{2804} (words related to seriousness and urgency) \\ \cmidrule(lr){1-3}
		0.537 & \nplink{8}{19141} (phrases related to legal or governmental mandates) & \nplink{9}{620} (references to political or legal mandates) \\ \cmidrule(lr){1-3}
		0.568 & \nplink{6}{19445} (phrases related to improving or reinforcing various aspects, such as control, security, ties, and relations) & \nplink{7}{10759} (phrases related to strengthening or reinforcement) \\ \cmidrule(lr){1-3}
		0.667 & \nplink{2}{22884} (references to the name \enquote{Sig} in various contexts) & \nplink{3}{14314} (names or terms related to the name \enquote{Sig}) \\ \cmidrule(lr){1-3}
		0.622 & \nplink{6}{3954} (expressions related to showing approval or disapproval) & \nplink{7}{18469} (mentions of thumbs used metaphorically or literally in various contexts) \\ \cmidrule(lr){1-3}
		0.758 & \nplink{4}{23164} (phrases related to monitoring or paying attention to specific things or people) & \nplink{5}{6236} (phrases instructing to pay attention to specific things or events) \\ \cmidrule(lr){1-3}
		0.750 & \nplink{9}{21027} (dates in the format of “day number ordinal indicator month year”) & \nplink{10}{21989} (dates expressed in numerical format) \\ \cmidrule(lr){1-3}
		0.898 & \nplink{10}{22616} (adjectives that describe the intensity of something) & \nplink{11}{16899} (exaggeratedly positive or negative sentiments) \\ \cmidrule(lr){1-3}
		0.834 & \nplink{7}{21991} (references to intellectual property such as patents and trademarks) & \nplink{8}{16537} (phrases related to patents and trademarks) \\ \cmidrule(lr){1-3}
		0.916 & \nplink{4}{8882} (terms related to financial derivatives) & \nplink{5}{12890} (terms related to financial derivatives) \\ \cmidrule(lr){1-3}
		0.982 & \nplink{4}{23820} (questions or debates revolving around specific topics) & \nplink{5}{22527} (phrases indicating doubt or uncertainty) \\
		\bottomrule
	\end{tabularx}
	\caption{Explanation pairs for different values of Jaccard}
	\label{tbl:jaccard-explanations}
\end{table}

\begin{table}[htp]
	\begin{tabularx}{\textwidth}{Y{0.1}Y{0.4}Y{0.4}}
	\toprule
	\textbf{Similarity} & \textbf{Upstream feature} & \textbf{Downstream feature} \\
	\midrule
		0.000 & \nplink{8}{15979} (mentions of open letters or discussions) & \nplink{9}{9352} (mentions of dense or concentrated things, such as thick fog or thick forests) \\ \cmidrule(lr){1-3}
		0.000 & \nplink{2}{2607} (words related to or containing the string \enquote{irk}) & \nplink{3}{1435} (words related to the concept of defense in various contexts) \\ \cmidrule(lr){1-3}
		0.141 & \nplink{9}{6305} (words related to mapping or plotting) & \nplink{10}{17644} (words related to physical body parts and violent actions) \\ \cmidrule(lr){1-3}
		0.160 & \nplink{10}{20113} (names of notable individuals and organizations) & \nplink{11}{10398} (references and information related to conflict and military actions in Yemen) \\ \cmidrule(lr){1-3}
		0.227 & \nplink{1}{15118} (phrases related to inhibition and inhibitors) & \nplink{2}{19226} (words related to scientific and technical terms like \enquote{baseline,} \enquote{experiment,} and \enquote{equilibrium.}) \\ \cmidrule(lr){1-3}
		0.333 & \nplink{2}{8570} ( mentions of different technology platforms and devices, specifically iOS and Android) & \nplink{3}{14919} (locations or geographical names) \\ \cmidrule(lr){1-3}
		0.435 & \nplink{10}{11796} (references to specific time periods, such as centuries and years) & \nplink{11}{19839} (phrases related to conclusions or summations) \\ \cmidrule(lr){1-3}
		0.498 & \nplink{0}{22953} (phrases related to food preparation) & \nplink{1}{4507} (phrases related to preparation or readiness) \\ \cmidrule(lr){1-3}
		0.571 & \nplink{2}{10918} (technical terms related to computer programming) & \nplink{3}{2511} (words related to uproar, chaos, and controversy) \\ \cmidrule(lr){1-3}
		0.507 & \nplink{6}{8639} (data transfer rates, storage capacities, and other technical specifications) & \nplink{7}{8346} (technical data related to electricity production and consumption) \\ \cmidrule(lr){1-3}
		0.634 & \nplink{0}{5889} (terms related to a specific type of technology or material, potentially related to chemical processes) & \nplink{1}{17831} (names of places or countries) \\ \cmidrule(lr){1-3}
		0.642 & \nplink{2}{2249} (currency amounts) & \nplink{3}{21826} (dollar amounts mentioned in a sentence) \\ \cmidrule(lr){1-3}
		0.787 & \nplink{5}{2675} (instances in which the word \enquote{it} is emphasized) & \nplink{6}{10500} (phrases containing \enquote{it is}) \\ \cmidrule(lr){1-3}
		0.776 & \nplink{8}{57} (statements or assertions) & \nplink{9}{14981} (phrases indicating the consequences or implications of a particular situation) \\ \cmidrule(lr){1-3}
		0.826 & \nplink{6}{6538} (exact descriptions or comparisons) & \nplink{7}{24347} (specific instances of an exact match in text, such as a repeated phrase or concept) \\ \cmidrule(lr){1-3}
		0.812 & \nplink{6}{15935} (adjectives and verbs indicating change or progress) & \nplink{7}{19867} (comparisons or statements about the state or change of things) \\ \cmidrule(lr){1-3}
		0.919 & \nplink{3}{22147} (phrases indicating a request or call for action) & \nplink{4}{5847} (phrases indicating an action or decision to pursue a particular course of action) \\ \cmidrule(lr){1-3}
		0.978 & \nplink{2}{13173} (the word \enquote{wild} and words associated with wildness or untamed nature) & \nplink{3}{10500} (words related to the concept of unpredictability or chaos) \\
		\bottomrule
	\end{tabularx}
	\caption{Explanation pairs for different values of necessity}
	\label{tbl:necessity-explanations}
\end{table}

\begin{table}[htp]
	\begin{tabularx}{\textwidth}{Y{0.1}Y{0.4}Y{0.4}}
	\toprule
	\textbf{Similarity} & \textbf{Upstream feature} & \textbf{Downstream feature} \\
	\midrule
		0.000 & \nplink{7}{20447} (mentions of the concept of democracy) & \nplink{8}{21210} (superlatives and adjectives suggesting exceptional qualities or rankings) \\ \cmidrule(lr){1-3}
		0.000 & \nplink{10}{15422} (mentions of physical cleaning actions, especially involving hands and face) & \nplink{11}{4381} (text related to investigations or examining matters) \\ \cmidrule(lr){1-3}
		0.129 & \nplink{0}{932} (mentions of hats) & \nplink{1}{5807} (university names and academic terms) \\ \cmidrule(lr){1-3}
		0.288 & \nplink{9}{22155} (phrases related to strong opinions or beliefs) & \nplink{10}{1972} (the word \enquote{thought} followed by a number, indicating contemplation or consideration) \\ \cmidrule(lr){1-3}
		0.286 & \nplink{4}{3091} (the name \enquote{Carmelo} along with various other related words and numbers) & \nplink{5}{23292} (events or entities related to specific \enquote{Expos} and person names, such as Neo and Maya) \\ \cmidrule(lr){1-3}
		0.308 & \nplink{0}{4369} (names or references to the basketball player Carmelo Anthony) & \nplink{1}{9680} (geographical locations) \\ \cmidrule(lr){1-3}
		0.310 & \nplink{0}{761} (occurrences of words related to 'total' or 'summing up') & \nplink{1}{19865} (universities and colleges names) \\ \cmidrule(lr){1-3}
		0.458 & \nplink{1}{16591} (words related to administrative or governmental entities, specifically the term \enquote{cant} or \enquote{cantonment}) & \nplink{2}{18040} (mentions or references to the word \enquote{Cantonese.}) \\ \cmidrule(lr){1-3}
		0.579 & \nplink{2}{9833} (mentions of specific brands, particularly HTC) & \nplink{3}{22363} (technology-related terms, particularly related to specific computer brands and components) \\ \cmidrule(lr){1-3}
		0.586 & \nplink{9}{11886} (phrases related to academic or professional affiliations, particularly at institutions or organizations) & \nplink{10}{19686} (information related to economic or financial reports and studies) \\ \cmidrule(lr){1-3}
		0.628 & \nplink{8}{21608} (mentions of prayer and religious belief, particularly related to Islam) & \nplink{9}{3951} (phrases related to computer programming and code) \\ \cmidrule(lr){1-3}
		0.606 & \nplink{1}{23792} (words related to appropriation and misappropriation) & \nplink{2}{4784} (keywords related to legality, propriety, and prudence) \\ \cmidrule(lr){1-3}
		0.746 & \nplink{8}{19540} (information and references related to ancient civilizations) & \nplink{9}{21368} (mentions of ancient civilizations, cultures, and history) \\ \cmidrule(lr){1-3}
		0.792 & \nplink{9}{19248} (criticisms and negative sentiments in the text) & \nplink{10}{23056} (issues or complexities within processes or environments) \\ \cmidrule(lr){1-3}
		0.846 & \nplink{5}{20436} (locations or places, particularly beaches) & \nplink{6}{22365} (references to a specific geographical location, in this case, \enquote{Palm Beach.}) \\ \cmidrule(lr){1-3}
		0.832 & \nplink{7}{878} (references to achieving success or winning in various contexts) & \nplink{8}{4743} (phrases related to success or achievement) \\ \cmidrule(lr){1-3}
		0.961 & \nplink{6}{22811} (mentions of geographic locations) & \nplink{7}{2432} (phrases related to states or conditions) \\ \cmidrule(lr){1-3}
		0.963 & \nplink{1}{19862} (phrases related to issues or actions involving workers) & \nplink{2}{22786} (references to workers in various contexts, such as productivity, accidents, and disputes) \\
		\bottomrule
	\end{tabularx}
	\caption{Explanation pairs for different values of sufficiency}
	\label{tbl:sufficiency-explanations}
\end{table}

\FloatBarrier
\section{Pass-through features at different thresholds} \label{app:pass-through}
\begin{figure}[htp]
    \centering
    \includegraphics[width=0.6\textwidth]{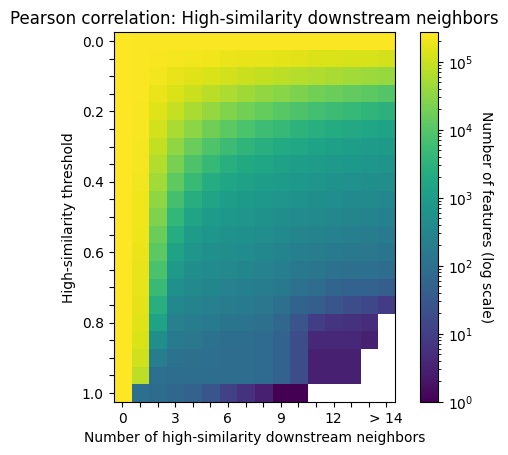}
    \caption{Number of high-similarity downstream neighbours for Pearson as a function of the high-similarity threshold.}
    \label{fig:pearson-neighbours}
\end{figure}

The number of pass-through vs. appearing and dying features is determined by the similarity threshold defining whether a feature has "passed-through" to the next layer or has no counterpart and has "disappeared". \cref{fig:pearson-neighbours} shows the number of high-similarity upstream and downstream neighbours (across all layer pairs) as a function of the similarity threshold. As expected, there are virtually no passed-through features when the threshold is set to 1, and to get a substantial number of features with more than 10 high-similarity neighbours, we have to choose a threshold of 0.2 or lower.

To find a suitable threshold, i.e., to calibrate the threshold to separate equivalent features from mere similar ones, we run the following interactive experiment:
\begin{enumerate}
    \item Start with a threshold of 0.5 and create feature pairs whose similarity is close to this threshold
    \item Show random pairs from this set, together with their GPT-3.5 explanations
    \item Experimenter decides whether the features of each pair seem to have equivalent meanings
    \item Depending on the answer, adjust the threshold to perform a binary search
    \item Stop when the threshold range is sufficiently narrow
\end{enumerate}

Running this experiment, we found that $\approx 0.95$ is a suitable threshold for Pearson.
\FloatBarrier %

\section{Logic gates} \label{app:logic-gates}
The argument from \cref{sec:results:logic-gates} that a high value of $n(f_u, f_d)$ means that \enquote{$f_d$ approximately implies $f_u$} equally applies to the \textit{sufficiency} measure, where $s(f_u, f_d)$ means that \enquote{$f_u$ approximately implies $f_d$}. By finding a feature $f_d$ with multiple high-similarity upstream neighbours $f_u^{(1)}, ..., f_u^{(n)}$, as done in \cref{tbl:logic-gates}, we can systematically find approximate \texttt{AND} relationships $f_d = f_u^{(1)} \wedge ... \wedge f_u^{(n)}$, where the meaning (i.e., the explanation) of $f_d$ is the intersection of the meanings of $f_u^{(n)}$ through $f_u^{(n)}$. Similarly, \texttt{OR} relationships are found by the sufficiency measure. However, it is important to note that as the number of neighbours increases, the accuracy of these relationships tends to diminish rapidly due to the potential compounding of errors.

The Pearson and Jaccard measures are less clear in what dependency they identify between the features' activation patterns: Pearson extracts the strength of the best linear relationship, while Jaccard measures the overlap between the binarized activation patterns. 

In \cref{tbl:app:logic-gates}, we compare our four similarity measures with respect to the informative value of this type of connection: For each measure, we extract feature groups consisting of a single downstream feature and exactly two high-similarity upstream neighbours. We then add explanations to each feature and manually gauge the interpretability of the link between the explanations. Our observations match the theoretical prediction that \emph{necessity} and \emph{sufficiency} and more interpretable in terms of implications than \emph{Pearson} and \emph{Jaccard}.

\begin{table}[htp]
    \begin{tabularx}{\textwidth}{Y{0.1}Y{0.58}Y{0.225}}
        \toprule
        \textbf{Measure} & \textbf{Upstream features} & \textbf{Downstream feature} \\
        \midrule
        Necessity &
        \begin{compactitemize}
            \item \nplink{1/3696} (numerical values corresponding to episodes in a series)
            \item \nplink{1/9152} (references to specific episodes)
        \end{compactitemize} &
        \nplink{2}{9633} (references to specific episodes in a TV series) \\
        \cmidrule(lr){1-3}
        Necessity &
        \begin{compactitemize}
            \item \nplink{0}{22068} ( words related to redesigning or altering something)
            \item \nplink{0}{24540} (words related to construction and building refurbishment)
        \end{compactitemize} &
        \nplink{1}{8399} (mentions of renovation or redevelopment of buildings or sites) \\
        \cmidrule(lr){1-3}
        Sufficiency &
        \begin{compactitemize}
            \item \nplink{3}{8555} (legal or financial terms related to seeking some sort of action or outcome)
            \item \nplink{3}{13751} (the word \enquote{sought})
        \end{compactitemize} &
        \nplink{4}{12885} (words related to seeking or looking for something) \\
        \cmidrule(lr){1-3}
        Sufficiency &
        \begin{compactitemize}
            \item \nplink{0}{17409} (terms related to internet browsers - both desktop and mobile)
            \item \nplink{0}{17478} (mentions of web browsers or actions related to web browsing)
        \end{compactitemize} &
        \nplink{1}{15150} (terms related to web browsers) \\
        \cmidrule(lr){1-3}
        Pearson &
        \begin{compactitemize}
            \item \nplink{0}{19702} (contractions of words to identify emphasis, assertion, or intent in text)
            \item \nplink{0}{23878} (words related to completion or fulfillment)
        \end{compactitemize} &
        \nplink{1}{12702} (phrases related to inviting and sharing information in a formal context) \\
        \cmidrule(lr){1-3}
        Pearson &
        \begin{compactitemize}
            \item \nplink{0}{17186} (instances of time-related phrases, particularly mentioning the number of days)
            \item \nplink{0}{21359} (words related to time, specifically focusing on periods of days)
        \end{compactitemize} &
        \nplink{1}{11771} (phrases related to the concept of time durations) \\
        \cmidrule(lr){1-3}
        Jaccard &
        \begin{compactitemize}
            \item \nplink{0}{5985} (concepts related to interpersonal relationships involving exploitation or taking advantage of others)
            \item \nplink{0}{20876} (phrases from the context repre- senting opinions or perspectives from others)
        \end{compactitemize} &
        \nplink{1}{18220} (instances where actions are performed on or with the involvement of others) \\
        \cmidrule(lr){1-3}
        Jaccard &
        \begin{compactitemize}
            \item \nplink{0}{16222} (political party names, especially the Republican Party)
            \item \nplink{0}{20665} (phrases related to events, possibly including the word \enquote{party})
        \end{compactitemize} &
        \nplink{1}{2979} (references to the Communist Party) \\
        \bottomrule
    \end{tabularx}
    \caption{Comparison of logic gate identification using all similarity measures, filtering for features with two high-similarity neighbours.}
    \label{tbl:app:logic-gates}
\end{table}

Without defining a quantitative measure, manual inspection indicates that the two \enquote{one-directional implication measures}, necessity and sufficiency, result in more straightforward semantic connections between features. Of course, this can extended to features with more than two high-similarity upstream neighbours, but manual interpretation becomes harder with more upstream connections.

\section{Relationship between ablation effect and similarity measure} \label{app:ablation}
To compute ablation effects, we ran the following algorithm, following the example set by \citep{marks-2024} to ensure that the reconstruction loss of the SAE at layer $k$ didn't affect the value of any of the features in layer $k+1$
\begin{enumerate}
    \item Select a feature $f_i$ in layer $k$'s residual stream SAE.
    \item Run a batch of tokens through the model.
    \item At layer $k$, cache activations across the batch for every SAE feature $f_j$ in layer $k+1$, and cache the reconstruction error (difference between original residual stream and residual stream reconstructed from SAE activations).
    \item Run the same batch through the model.
    \item Intervene in the model at layer $k$:
    \begin{itemize}
        \item Encode residual stream into SAE.
        \item Ablate $f_i$, i.e. manually set it to 0.
        \item Reconstruct residual stream by decoding the ablated SAE feature vector.
        \item Add back the cached reconstruction error stored in step 3.
    \end{itemize}
    \item Pass this modified residual stream to layer $k+1$, cache downstream SAE feature values
    \item Calculate absolute value of the mean difference in $f_j$ activations for each token with/without ablating $f_i$
\end{enumerate}

Given that this cannot be run in parallel for multiple upstream features (since ablating two features simultaneously could create interference), we needed to be judicious about which upstream features we chose to ablate.
For each similarity measure, for each layer pair, we selected 10 evenly-sized bins of similarity measure scores and randomly 10 sampled feature pairs per each layer pair (e.g. 10 pairs with Pearson between 0.1 and 0.2), and then ablated the upstream feature and recorded the downstream effect.

\begin{figure}[H]
    \centering
    \includegraphics[width=0.8\textwidth]{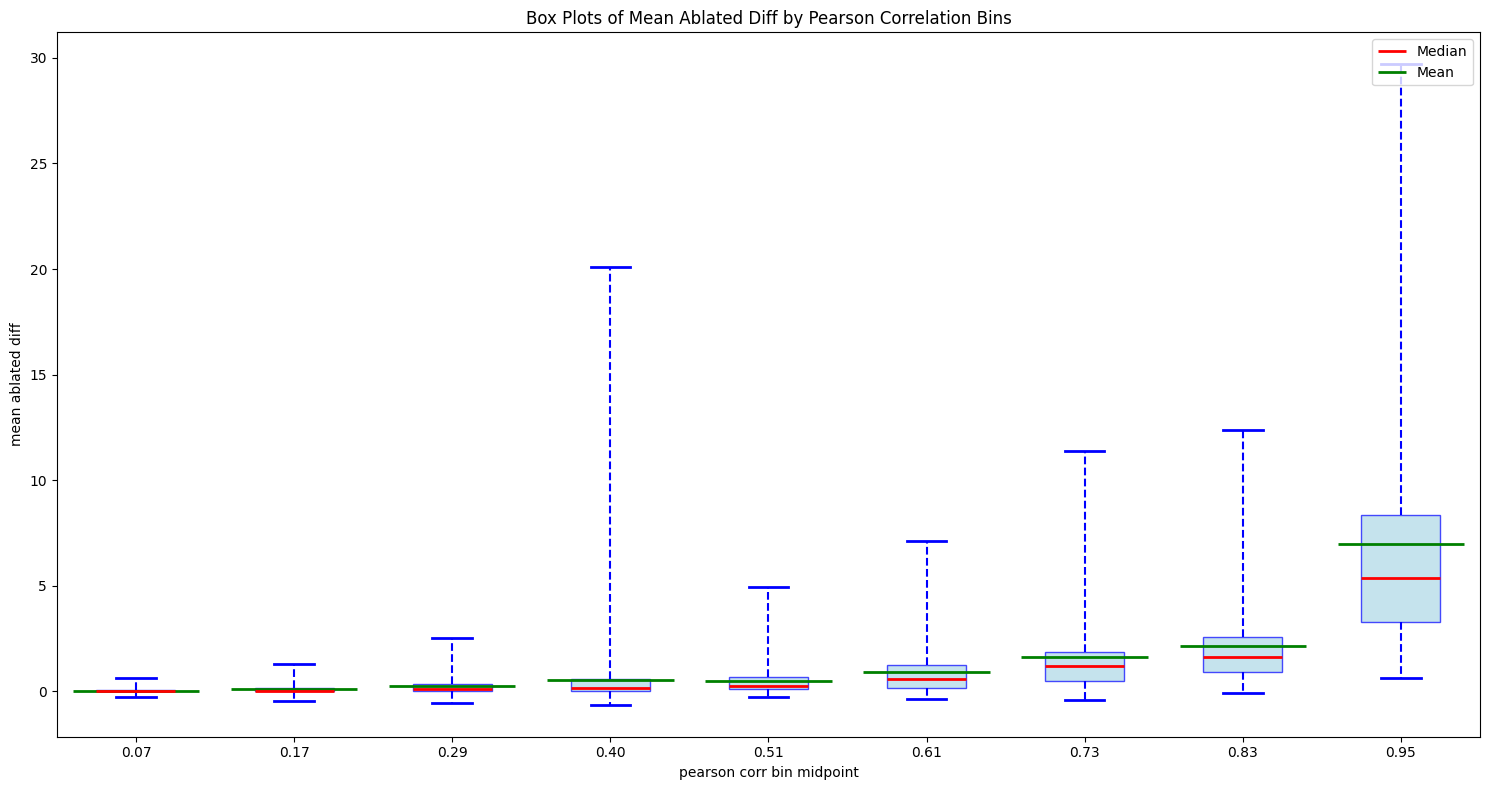}
    \caption{}
    \label{fig:pearson-boxplot}
\end{figure}

\begin{figure}[H]
    \centering
    \includegraphics[width=0.8\textwidth]{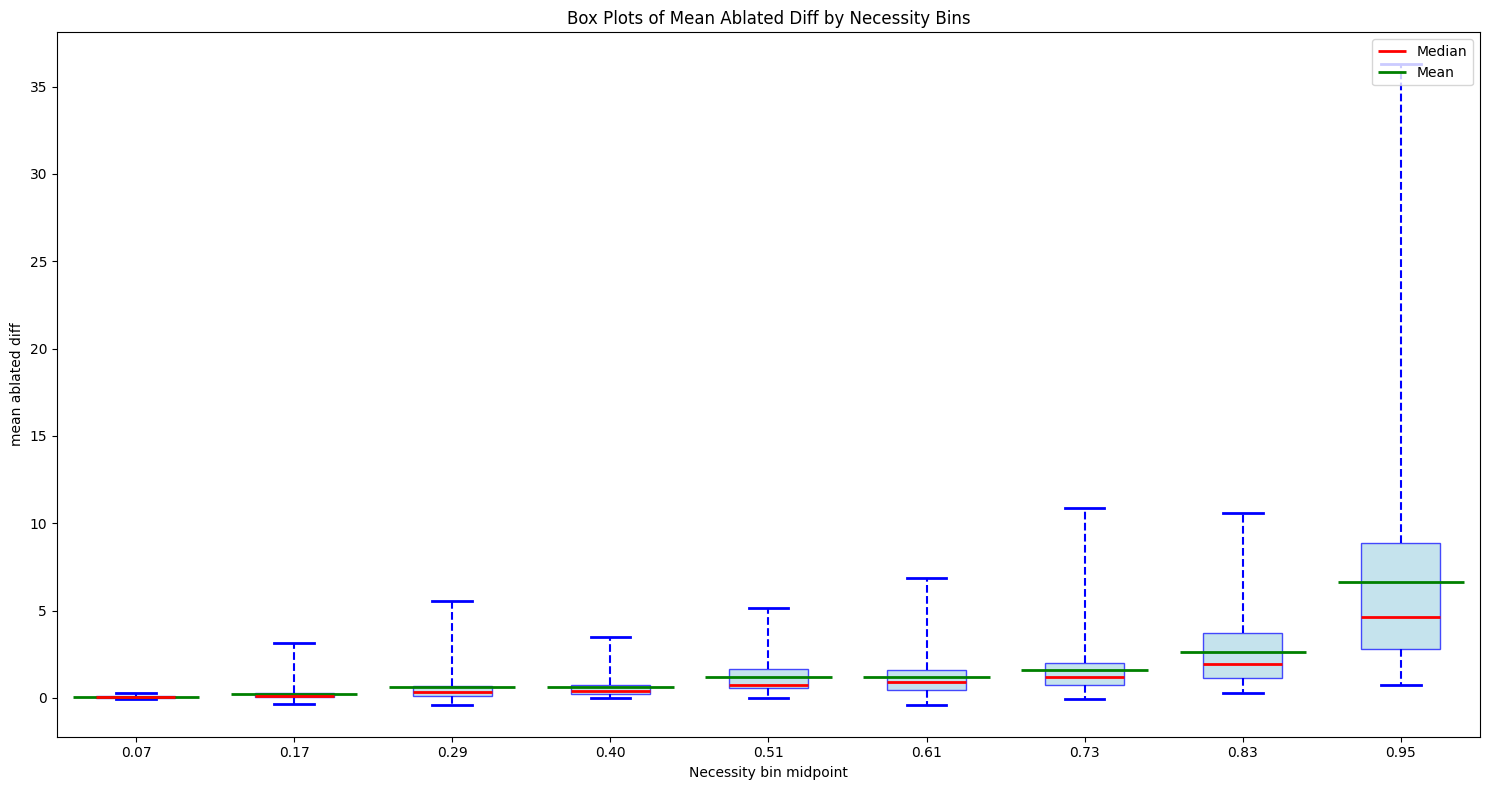}
    \caption{}
    \label{fig:necessity-boxplot}
\end{figure}

\begin{figure}[H]
    \centering
    \includegraphics[width=0.8\textwidth]{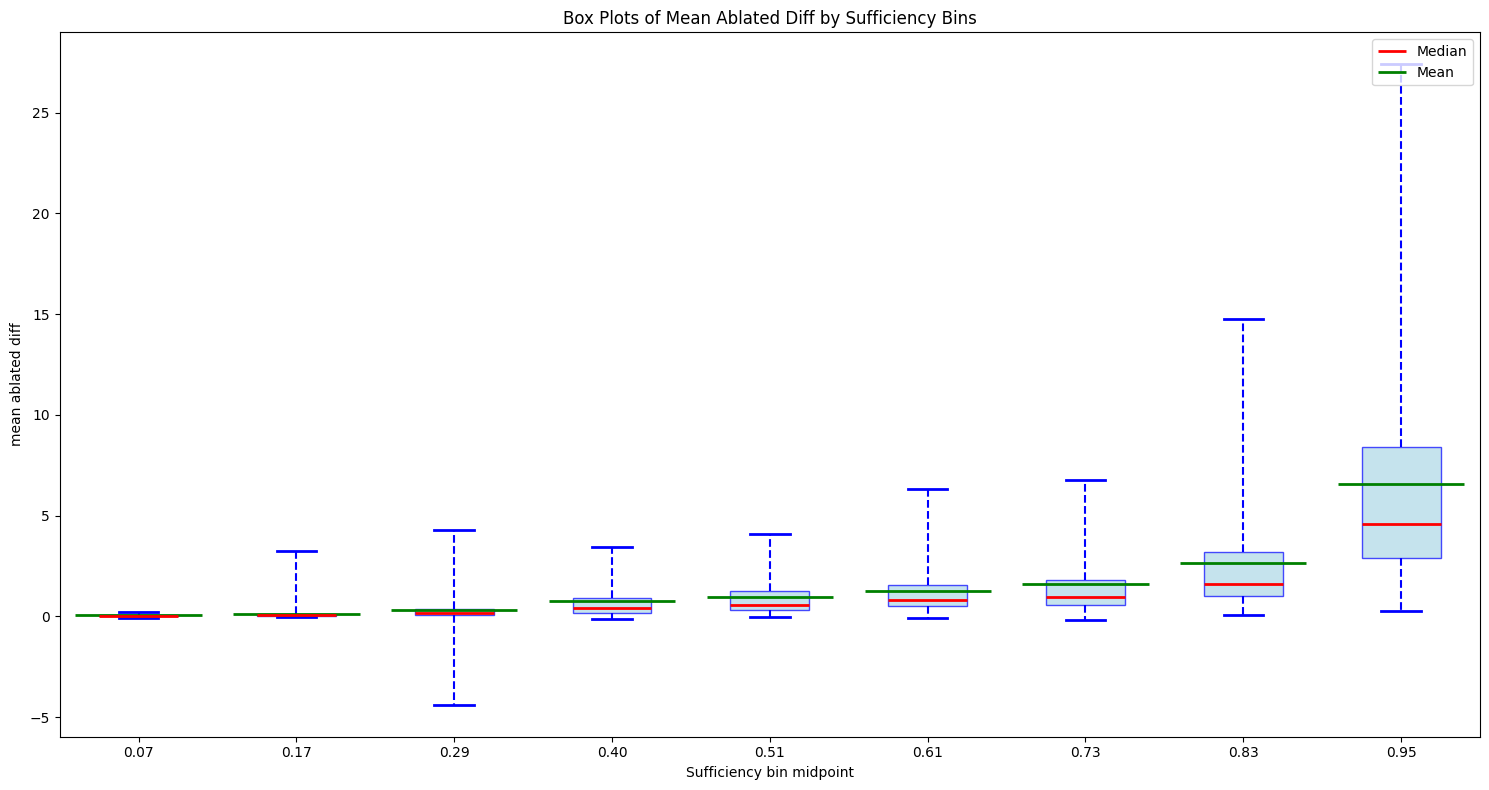}
    \caption{}
    \label{fig:sufficiency-boxplot}
\end{figure}

\begin{figure}[H]
    \centering
    \includegraphics[width=0.8\textwidth]{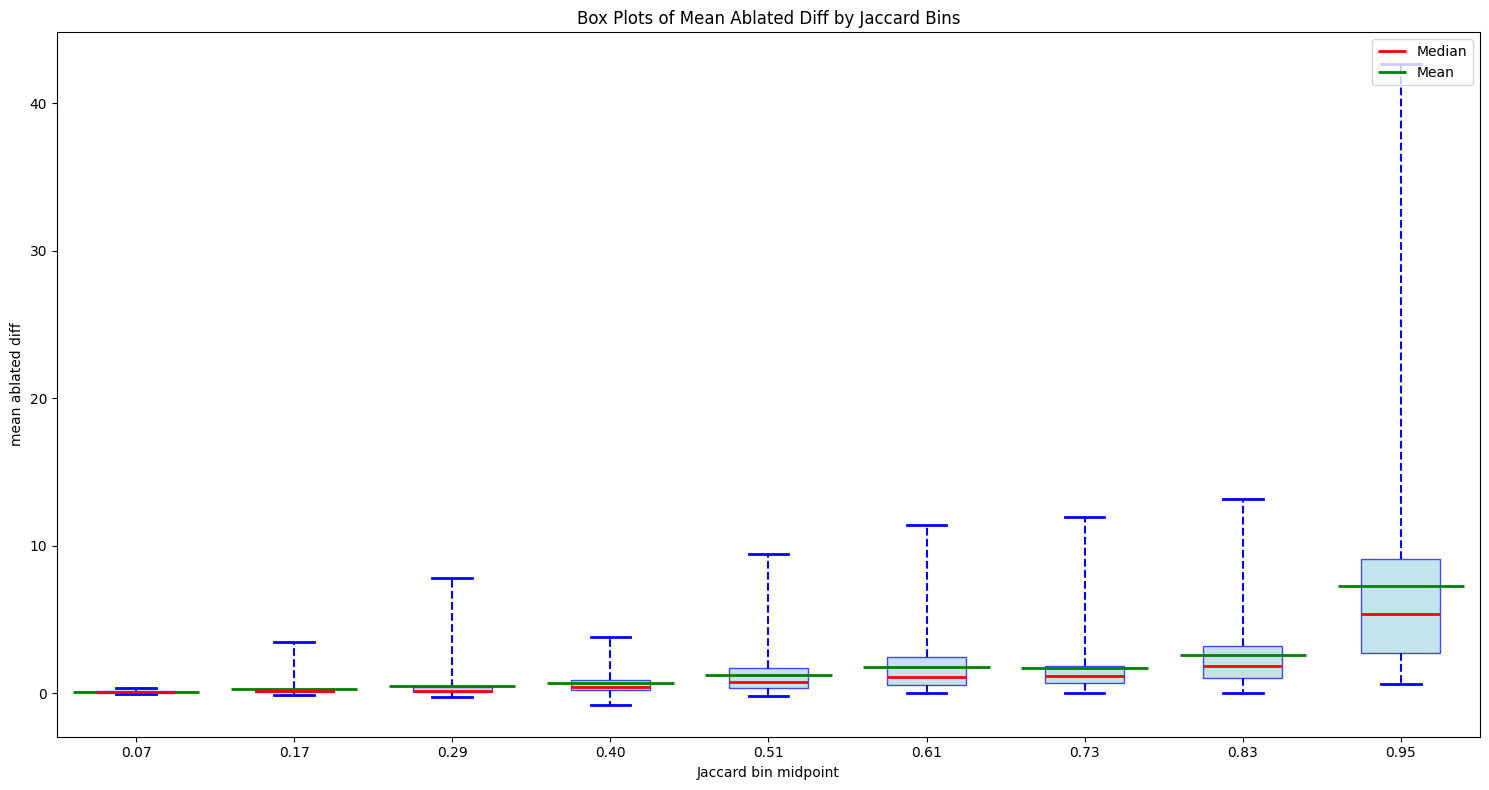}
    \caption{}
    \label{fig:jaccard-boxplot}
\end{figure}

As described above, each similarity measure bin had 10 samples per each of the 11 layer pairs, and the corresponding absolute value of the average change in downstream activation was plotted with boxplots.
The mean ablation effect increases for higher interaction values across all similarity measures, validating that those similarity measures are at least weak indicators of the true causal effect, although Jaccard has the weakest relationship.

A limitation of this approach is, given that we only ablated one feature at a time, any upstream feature redundantly causing a downstream feature's activation alongside a \enquote{sibling} upstream feature would appear to have no causation on the downstream feature (because if one is ablated, the sibling would still cause the downstream feature to fire, see \cite{mueller-2024}).
\FloatBarrier

\section{Error projection method}
\label{app:projectionmethod}
\begin{enumerate}
    \item Feed 10M tokens through the model and consider whenever an SAE feature's activation in any layer hit at least 10\% of its max activation
    \item Calculate the next-layer reconstruction error for that token by
    \begin{enumerate}
        \item Recording the residual stream at the next layer
        \item Encoding and decoding the next layer residual stream with the next layer SAE, the \enquote{reconstructed} residual stream
        \item Computing the difference between the original residual stream and the reconstructed residual stream i.e. where x is the next layer residual stream, computing $\varepsilon$ in  $x= \hat{x} + \varepsilon(x) = (xW_{enc}+b_{enc})W_{dec}+b_{dec}+\varepsilon(x)$ 
    \end{enumerate}
    \item Project that error term back onto the $W_{dec}$ direction of the previous layer's SAE for the feature that activated
\end{enumerate}

Then we plot, for each token, the activation against the error term.

\section{Community Detection}
\subsection{Graph Construction} \label{app:graph-construction}
There are myriad hyperparameters to consider when constructing an SAE graph for community detection, such as whether to use weighted or unweighted edges, which quality function to use and the edge weight threshold. One could construct a \enquote{full} graph where every node at layer $L$ is connected to every node at layer $L+1$, with an edge weight equal to the strength of the connection, i.e. the similarity measure. Alternatively, one could construct an unweighted graph with edges added if the similarity measure between them exceeds a threshold value, which therein adds another parameter affecting the resultant graph structure. The choice of graph construction impacts the method of community detection. As an example, both Louvain and Leiden community detection algorithms can be set to either consider edge weights or ignore them.

\subsection{Further Examples of Communities}

\begin{figure}[htbp]
    \centering
        \includegraphics[width=\textwidth]{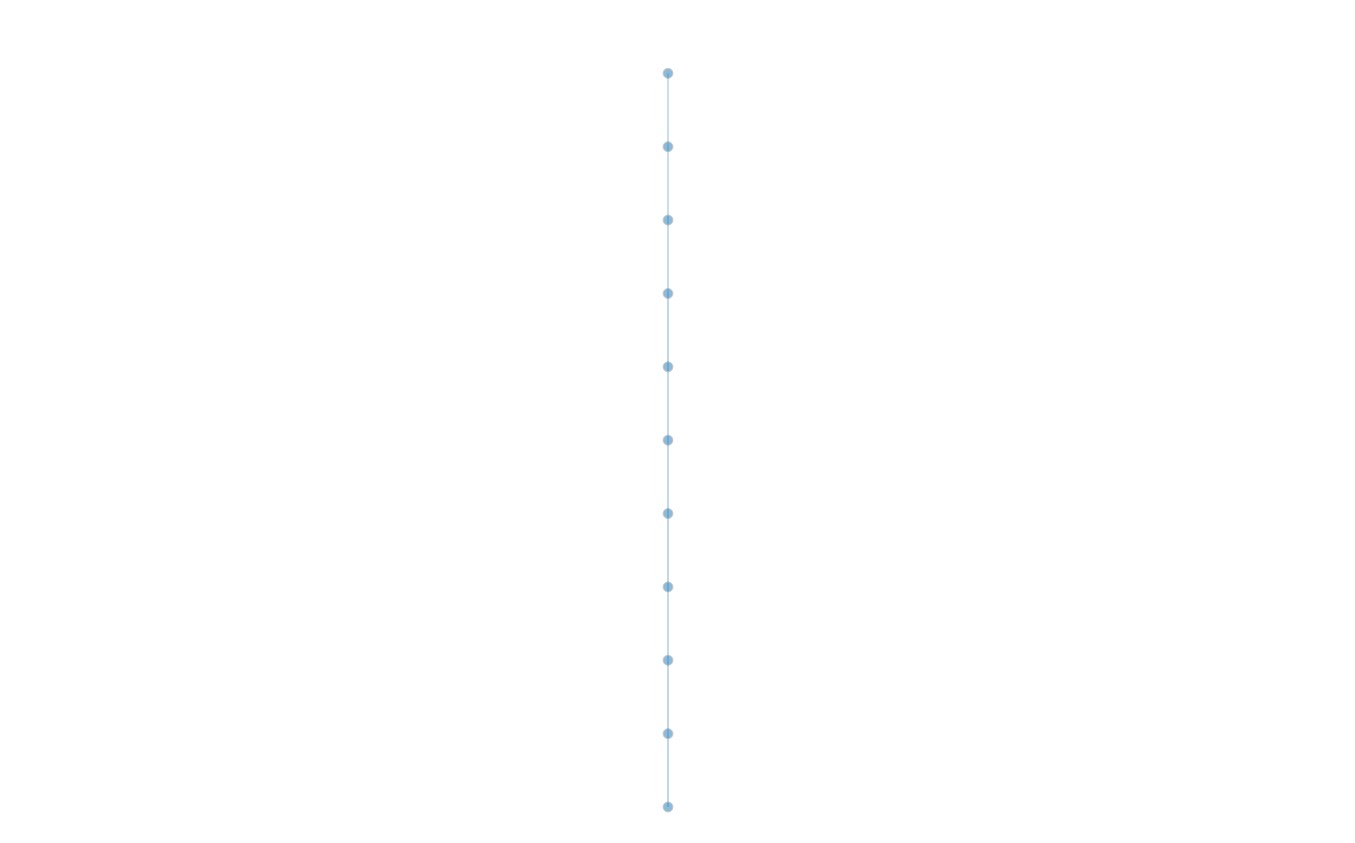}
        \caption{A community found by applying the Louvain algorithm on a graph with edge weights corresponding to Jaccard similarity. This community corresponds to the concept of the C++ programming language. Can be found with name: \texttt{jaccard\_similarity\_louvain\_threshold\_0.1\_size\_11\_1229} in the feature browser.}
    \label{fig:figures/jaccard-communities-louvain}
\end{figure}

\begin{figure}[htbp]
    \centering
        \includegraphics[width=\textwidth]{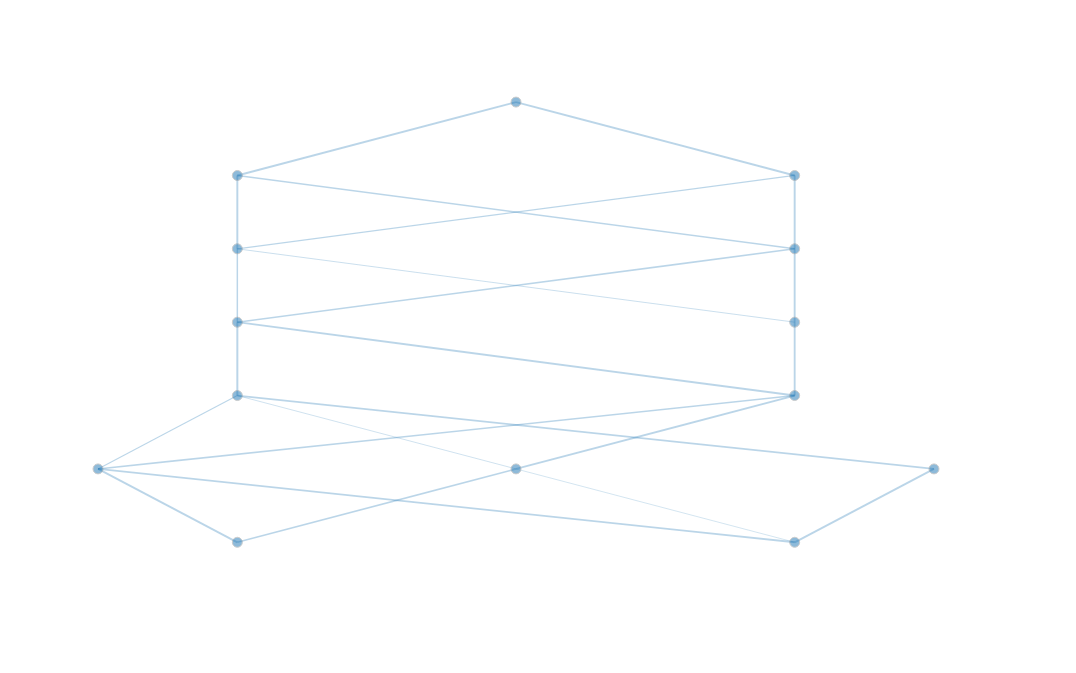}
        \caption{A community found by applying the Leiden algorithm with a modularity quality function on a graph with edge weights corresponding to sufficiency similarity. This community corresponds to the concept of operations of various kinds. Can be found with name: \texttt{sufficiency\_leiden\_modularity\_threshold\_0.1\_size\_14\_61} in the feature browser.}
    \label{fig:sufficiency-communities-leiden-modularity}
\end{figure}

\begin{figure}[htbp]
    \centering
        \includegraphics[width=\textwidth]{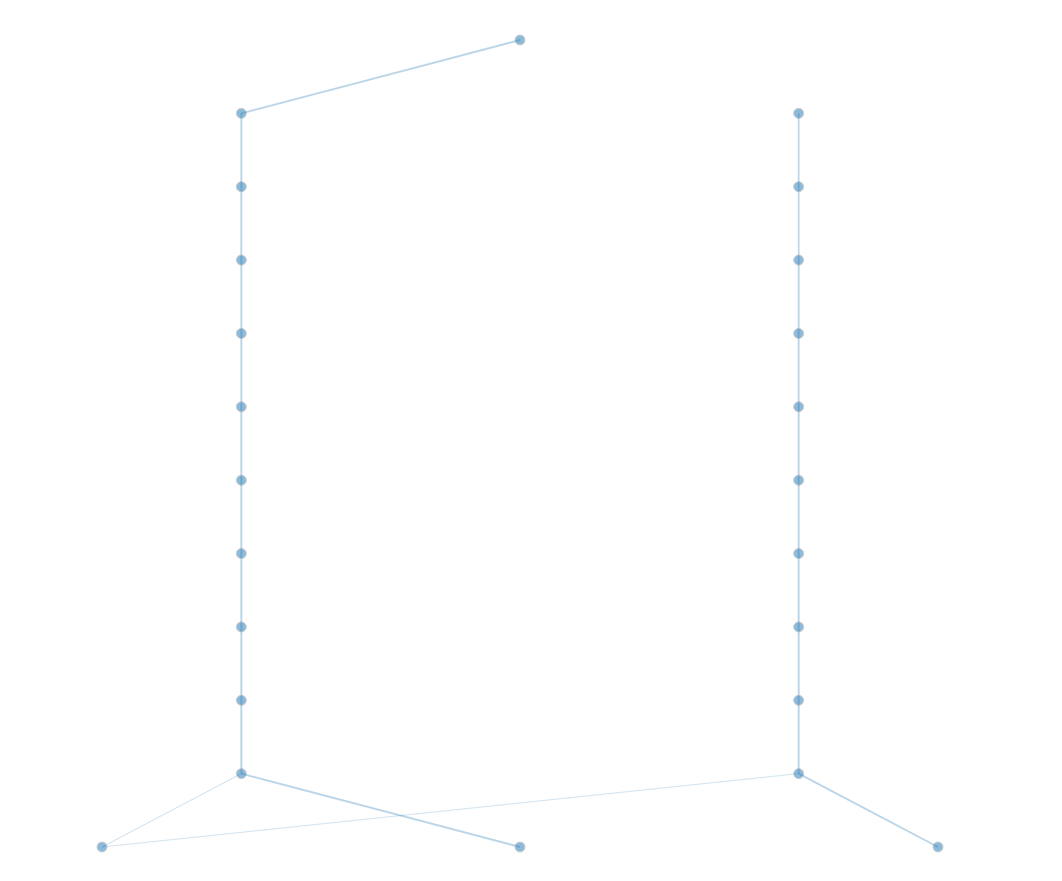}
        \caption{A community found by applying the Louvain algorithm on a graph with edge weights corresponding to Pearson similarity. This community corresponds to two concepts: the name "allen" and the name "adams". Can be found with name: \texttt{pearson\_louvain\_threshold\_0.1\_size\_24\_82} in the feature browser.}
    \label{fig:pearson-communities-louvain}
\end{figure}

\FloatBarrier
\section{Discrepancies between similarity scores and explanation similarity}
\label{app:autointerp-inaccuracies}

We find instances in which two features have a high similarity score, but the GPT-3.5 explanations of each are dissimilar. In many of these cases, after looking at the top-activating tokens manually, we find that one of explanations is inaccurate and that they should actually have very similar activations, because both features are detecting the same patterns in the input.

\begin{table}[htp]
    \begin{tabularx}{\textwidth}{XXp{1.5cm}}
        \toprule
        \textbf{Upstream feature with inaccurate summary} & \textbf{Downstream feature} & \textbf{Pearson score} \\
        \midrule
        \nplink{0}{60} (words related to padding or cushioning) - it's actually the words \enquote{pad} and \enquote{pads} &
        \nplink{1}{7834} (words related to objects with flat surfaces, often used for various purposes like writing, gaming, or medical applications) &
        0.9907  \\
        \cmidrule(lr){1-3}
        \nplink{0}{134} (words related to a specific geographical location, possibly a country or region) - it's actually \enquote{ria} &
        \nplink{1}{24445} (the term \enquote{ria} in various contexts) &
        0.992 \\
        \cmidrule(lr){1-3}
        \nplink{3}{7661} (names followed by a country or position) - it's actually \enquote{aur} &
        \nplink{4}{19195} (the word \enquote{aur}) &
        0.9976 \\
        \bottomrule
    \end{tabularx}
    \caption{Examples of inaccurate GPT3.5 summaries of features}
    \label{tbl:inaccurate-explanations}
\end{table}

\end{document}